\documentclass[preprint,12pt,authoryear]{elsarticle}

\usepackage{amssymb}
\usepackage{hyperref}
\usepackage{lineno}

\usepackage{algorithm}
\usepackage{algpseudocode}

\usepackage{amsmath} 

\usepackage{array}
\usepackage{todonotes}
\usepackage{natbib}
\usepackage{booktabs}

\usepackage{pdflscape} 
\usepackage{booktabs} 
\usepackage{graphicx} 
\usepackage{caption} 
\usepackage{geometry}
\usepackage{color}  

\journal{Elsevier}

\begin{document}

\begin{frontmatter}

\title{EvolveSignal: A Large Language Model Powered Coding Agent for Discovering Traffic Signal Control Strategies}


\author[a,f]{Leizhen Wang}
\author[a]{Peibo Duan\corref{cor1}}  
\author[b]{Hao Wang}
\author[c]{Yue Wang}
\author[d]{Jian Xu}
\author[e]{Nan Zheng}
\author[f]{Zhenliang Ma\corref{cor1}}  

\cortext[cor1]{Corresponding author}


\affiliation[a]{organization={Department of Data Science and Artificial Intelligence, Monash University},
            city={Melbourne},
            country={Australia}}

\affiliation[b]{organization={School of Transportation, Southeast University},
            city={Nanjing},
            country={People's Republic of China}}

\affiliation[c]{organization={R\&D Center, Zhejiang Dahua Technology CO.,  LTD},
            city={Hangzhou},
            country={People's Republic of China}}
            
\affiliation[d]{organization={Institute of Automation, Chinese Academy of Sciences},
            city={Beijing},
            country={People's Republic of China}}
            
\affiliation[e]{organization={Department of Civil and Environmental Engineering, Monash University},
            city={Melbourne},
            country={Australia}}

\affiliation[f]{organization={Department of Civil and Architectural Engineering, KTH Royal Institute of Technology},
            city={Stockholm},
            country={Sweden}}

\begin{abstract}

In traffic engineering, fixed-time traffic signal control remains widely used for its low cost, stability, and interpretability. However, its design relies on hand-crafted formulas (e.g., Webster) and manual re-timing by engineers to adapt to demand changes, which is labor-intensive and often yields suboptimal results under heterogeneous or congested conditions. This paper introduces \textbf{EvolveSignal}, an LLM-powered coding agent for automatically discovering interpretable heuristic strategies for fixed-time traffic signal control. Rather than deriving entirely new analytical formulations, the proposed framework focuses on exploring code-level variations of existing control logic and identifying effective combinations of heuristic modifications. We formulate the problem as program synthesis, where candidate strategies are represented as Python functions with fixed input–output structures and iteratively optimized through external evaluations (e.g., a traffic simulator) and evolutionary search. Experiments on a signalized intersection demonstrate that the discovered strategies outperform a classical baseline (Webster's method), reducing average delay by 20.1\% and average stops by 47.1\%. Beyond performance, ablation and incremental analyses reveal that EvolveSignal can identify meaningful modifications, such as adjusting cycle length bounds, incorporating right-turn demand, and rescaling green allocations, that provide useful insights for traffic engineers. This work highlights the potential of LLM-driven program synthesis for supporting interpretable and automated heuristic design in traffic signal control. The code for this paper is open-source and available at: \href{https://github.com/georgewanglz2019/EvolveSignal}{https://github.com/georgewanglz2019/EvolveSignal}.

\end{abstract}




\end{frontmatter}


\section{Introduction}
\label{sec:intro}

Traffic signal control remains the cost-effective strategy for mitigating urban congestion and improving traffic efficiency. In recent years, artificial intelligence (AI)-based methods have attracted significant attention, especially those utilizing reinforcement learning (RL) \citep{wei2019survey,wang2023human} and large language models (LLMs) \citep{movahedi2024crossroads,lai2025llmlight}. These data-driven approaches demonstrate the potential to learn high-performance control strategies that can enhance mobility at both the intersection and network levels.

However, most existing studies based on RL and LLMs focus on adaptive traffic signal control (ATSC), which dynamically adjusts signal phases and green times in response to real-time traffic conditions. Although effective, ATSC relies on extensive infrastructure, including high-resolution sensors (e.g., cameras, radar, inductive loops) and capable signal controllers with strong computational and communication resources. These requirements pose challenges in terms of cost, privacy, and deployment feasibility, limiting the scalability and practical implementation in real-world networks.

In contrast, fixed-time traffic signal control is still widely adopted due to its simplicity, low implementation cost, and operational stability \citep{zhou2024traffic,wang2025chat2spat}. More importantly, fixed-time control methods are inherently interpretable and transparent, making them easier to adjust and integrate with domain-specific rules. Optimizing fixed-time control plans remains a practically valuable and promising research direction, especially in scenarios where sensing and actuation resources are limited.

Traditional fixed-time control algorithms are typically derived from empirical engineering heuristics, supported by traffic flow theory, queuing models, and optimization methods. In this study, we define a \textit{fixed-time traffic signal control algorithm} as a method that computes a static signal timing plan based on low-resolution or aggregated traffic demand data, without requiring real-time sensing. The output typically includes a fixed cycle length and green time allocation. Among these, Webster's method \citep{webster1958traffic} is a classic example that uses critical flow ratios and cycle losses to derive closed-form solutions.

While these classical methods are highly interpretable and easy to implement, they are often insufficient to meet the specific operational needs of individual intersections. Due to heterogeneity in traffic demand, geometry, and performance objectives across intersections, a single globally applied strategy (e.g., directly using Webster's output) may lead to suboptimal outcomes. In practice, traffic engineers are frequently required to fine-tune or adapt these algorithms on a per-intersection basis by adjusting parameters, modifying constraints, or integrating additional domain rules to better match local conditions and achieve desired performance.

This leads to a critical research question: \textit{Can we move beyond hand-crafted rules and enable automated discovery of effective heuristic strategies for fixed-time control with higher adaptability and performance?}

To address this, we propose leveraging the code generation and reasoning capabilities of LLMs to explore variations of fixed-time control algorithms in executable form (e.g., Python functions). Rather than deriving entirely new analytical models, the goal is to automatically identify useful modifications and combinations of domain-inspired rules within an interpretable programming framework.

The main contributions of this paper are as follows:

\begin{itemize}
    \item We formulate fixed-time signal control as a \emph{program synthesis} problem, where an algorithm is represented as an executable function with a fixed input–output structure. The focus is on discovering interpretable heuristic structures that transform traffic features into signal timings.
    
    \item We develop \textbf{EvolveSignal}, an LLM-powered coding agent that explores and refines candidate programs through simulation-based evaluation and evolutionary search, enabling automated and auditable heuristic discovery.
    
    \item We validate EvolveSignal on synthetic traffic scenarios with heavy demand, showing improvements over a classical baseline while revealing interpretable modifications that provide useful insights for traffic engineering practice.
\end{itemize}

The remainder of this paper is organized as follows. Section~\ref{sec:review} reviews related work on conventional and AI-based traffic signal control. Section~\ref{sec:method} introduces the proposed \textbf{EvolveSignal} framework. Section~\ref{sec:experiment} presents the experimental setup, results, and ablation analyses. Section~\ref{sec:discussion} discusses insights and implications. Section~\ref{sec:conclusion} concludes the paper and outlines future directions.

\section{Related Work}
\label{sec:review}

\subsection{Conventional traffic signal control methods}
\label{sec:traditional_tsc}

Traditional traffic signal control methods for a single intersection can be divided into three groups. The first is fixed-time signal control, such as Webster's method, which optimizes green time using historical traffic data to minimize delay \citep{webster1958traffic}. The second is actuated signal control, where phases are adjusted based on real-time data from detectors \citep{fellendorf1994vissim}. Both methods, however, struggle to handle dynamic traffic patterns. A third approach, adaptive traffic signal control, dynamically adjusts phases and green times based on real-time conditions \citep{wang2018review,xie2025coordination}. 

From a control scope perspective, traditional methods can also be classified into single-intersection control, arterial coordination~\citep{gartner1991multi,wang2022coordinated}, and network-wide control~\citep{hunt1982scoot}. While these methods have a solid theoretical foundation and are relatively easy to implement, they often require the traffic management engineer to optimize each intersection individually based on local conditions, which is time-consuming, labor-intensive, and costly.

\subsection{AI-based traffic signal control methods}
\label{sec:AI_tsc}

Recently, RL-based approaches have shown significant promise in adaptive traffic signal control, achieving superior performance in terms of reduced delay \citep{chu2019multi,wang2023human,kuwahara2025optimum,jin2026hybrid,ren2025cooperative}, improved energy efficiency \citep{koch2023adaptive,peng2025combat}, and enhanced safety \citep{gong2020multi} compared to traditional methods. In parallel, LLM-based methods have shown unique advantages across various transportation domains \citep{ma2026large,nie2025exploring,guo2024towards,ling2026review,liu2025toward,wang2025agentic,pang2026large,qin2025foundational,zhen2025crashsage,qin2025lingotrip}.
In the context of adaptive traffic signal control, LLMs can deliver even better adaptive control performance while providing natural language reasoning and explanations\citep{movahedi2024crossroads,lai2025llmlight}. However, practical deployment remains challenging due to high costs, privacy concerns, and the complex infrastructure required to scale RL and LLM models. This study proposes a novel approach to better harness the potential of AI for traffic signal control, ensuring both scalability and real-world feasibility.

\section{Methodology}
\label{sec:method}

\subsection{Problem Definition}
\label{sec:problem}

\begin{figure}
  \centering
  \includegraphics[width=0.65\textwidth]{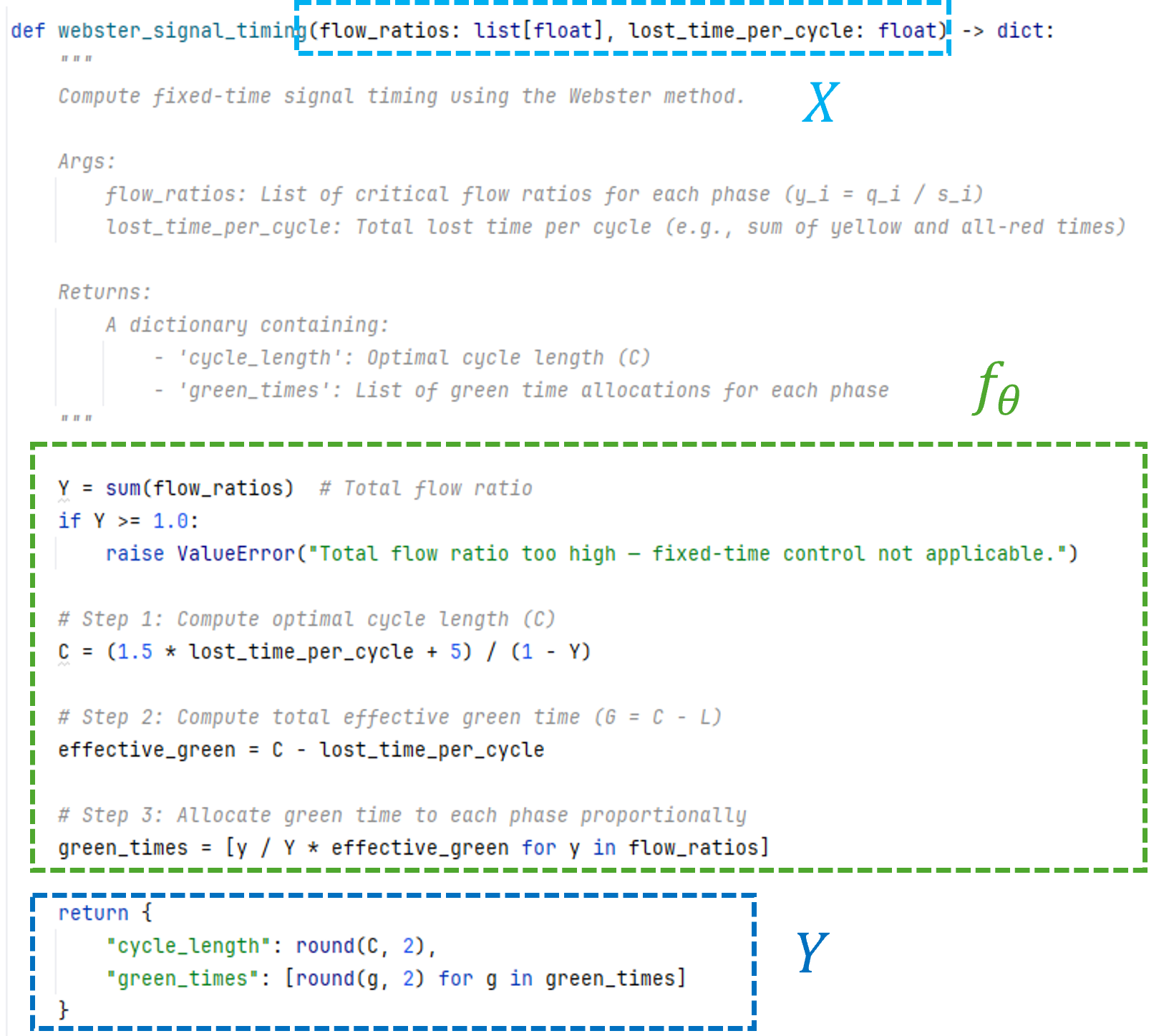}
  \caption{The problem definition in Python}
  \label{fig:problem}
\end{figure}

This study formulates the optimization of fixed-time traffic signal control algorithms as a program synthesis problem. The objective is to learn an executable function that maps intersection-specific traffic features to a complete signal timing plan. Formally, the problem is defined as:
\begin{equation}
    f_\theta: {X} \rightarrow {Y},
\end{equation}
where ${X}$ denotes the input space, including structural and traffic-related features of an intersection (e.g., traffic volume, geometric layout, and phase configuration); ${Y}$ represents the output space of fixed-time signal control parameters (e.g., cycle length, green time per phase); and $f_\theta$ is a parameterized executable program that encodes the logic and structure of the control algorithm using symbolic elements such as variables, operators, formula templates, and conditional statements, as shown in Figure~\ref{fig:problem}.

Unlike traditional analytical models based on pre-defined rules (e.g., Webster's formula), this study aims to automatically discover the structure of $f_\theta$ using a coding agent powered by LLMs and guided by evolutionary search. The LLM generates syntactically valid candidate programs in Python, while a simulation-based evaluation framework iteratively refines them based on traffic performance.

The optimization objective is to maximize the expected score over a distribution of traffic scenarios $\mathcal{D}$:
\begin{equation}
    \max_{f_\theta} \ \mathbb{E}_{x \sim \mathcal{D}} \left[ \text{Score}\left( f_\theta(x) \right) \right],
\end{equation}
where $\text{Score}(\cdot)$ is a task-specific evaluation metric that quantifies traffic efficiency, such as average throughput, delay reduction, or stop minimization.

This formulation enables the automatic discovery of interpretable, effective, and adaptable fixed-time signal control strategies, suitable for deployment in infrastructure-constrained environments.

\subsection{Overview of the Proposed Framework}
\label{sec:overview}

\begin{figure*}[ht]
  \centering
  \includegraphics[width=0.98\textwidth]{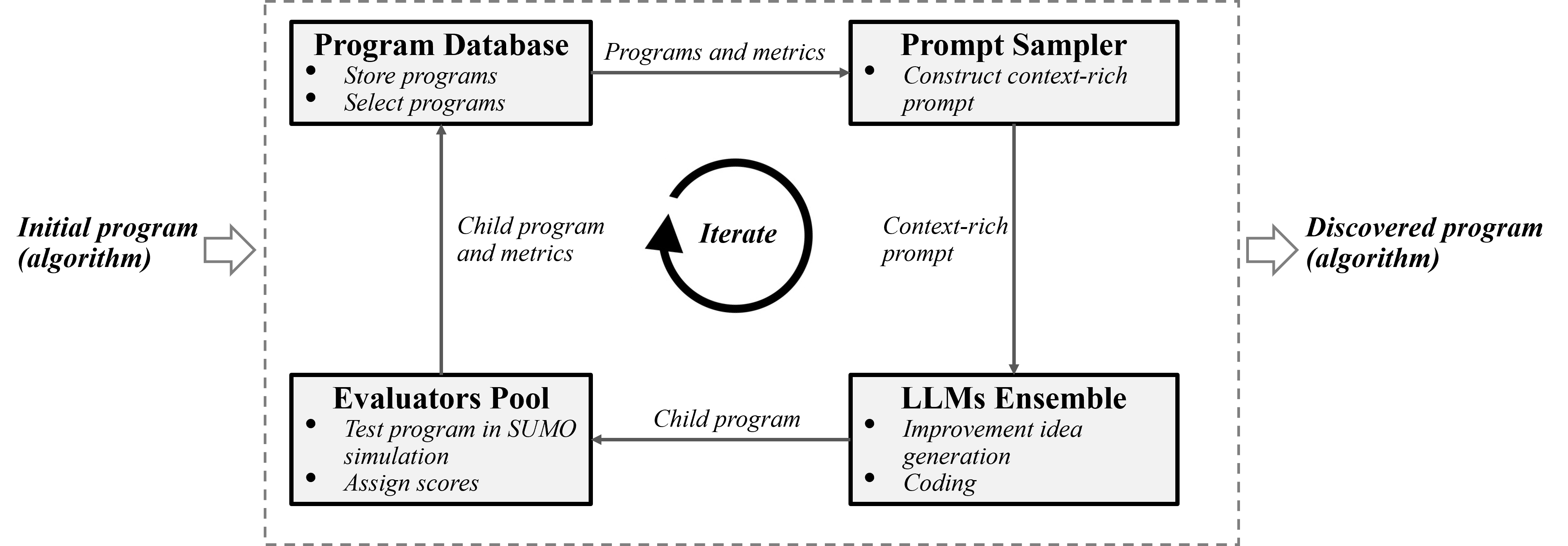}
  \caption{The EvolveSignal framework}
  \label{fig:framework}
\end{figure*}

Figure~\ref{fig:framework} shows the overall framework of \textbf{EvolveSignal}, an LLM-powered coding agent designed to evolve fixed-time traffic signal control algorithms. The process begins with an initial program (e.g., an implementation of Webster's method), which is iteratively modified, simulated, and evaluated until convergence. In each iteration, candidate programs are stored and organized in the \textit{Program Database}, which supports both performance tracking and diversity maintenance. A parent program and several inspiration programs are then passed to the \textit{Prompt Sampler}, which constructs prompts combining code, metrics, and search strategies. These prompts are processed by the \textit{LLMs Ensemble}, generating code-level modifications that act as crossover and mutation operators. The resulting child program is tested in the \textit{Evaluators Pool} using microscopic traffic simulation, and its performance is fed back to the Program Database. This loop continues until either the maximum number of iterations is reached or the best-performing algorithm is identified. Detailed descriptions of these modules are provided in Appendix~\ref{appendix:modules}.

\subsection{Pseudocode for the Proposed Framework}
\label{sec:pseudocode}

The workflow of EvolveSignal is summarized in Algorithm~\ref{alg:evolution}, adapted from the AlphaEvolve framework~\citep{novikov2025alphaevolve} with implementation modifications from the open-source OpenEvolve project~\citep{openevolve}. The pseudocode captures the evolutionary loop of selection, LLM-guided modification, simulation-based evaluation, and population update.

\begin{algorithm}[H]
\caption{Evolutionary Framework for Traffic Fixed-Time Signal Control}
\label{alg:evolution}
\begin{algorithmic}[1]
\State Implement initial program $f_{\theta_0}$ (e.g., Webster's method)
\State Initialize \textit{Program Database} ${P} \gets \{ f_{\theta_0} \}$
\State Initialize \textit{Evaluators Pool} for simulation
\For{$i = 1$ \textbf{to} $\text{max\_iterations}$}
    \State Sample parent program $f_\theta \in {P}$
    \State Select inspiration programs $\mathcal{I} \subset {P}$ based on top-k and diversity
    \State Construct prompt ${P}_{\text{prompt}}$ using $f_\theta$, $\mathcal{I}$, and performance metrics
    \State Sample an LLM model $G$ from \textit{LLMs Ensemble}
    \State $\text{modifications} = G({P}_{\text{prompt}})$
    \State Apply modifications to generate child program $f_\theta'$
    \State Evaluate $f_\theta'$ in SUMO~\citep{lopez2018microscopic}, compute performance score $\text{Score}(f_\theta')$
    \State Update \textit{Program Database} with $f_\theta'$ using MAP-Elites~\citep{mouret2015illuminating}
    \If{termination criteria satisfied} 
        \State \textbf{break}
    \EndIf
\EndFor
\State \Return Best discovered program $f_{\theta_{\text{best}}}$
\end{algorithmic}
\end{algorithm}

\section{Experiment}
\label{sec:experiment}

\subsection{Experimental Setup}
\label{sec:setting}

\begin{figure}
  \centering
  \includegraphics[width=0.4\textwidth]{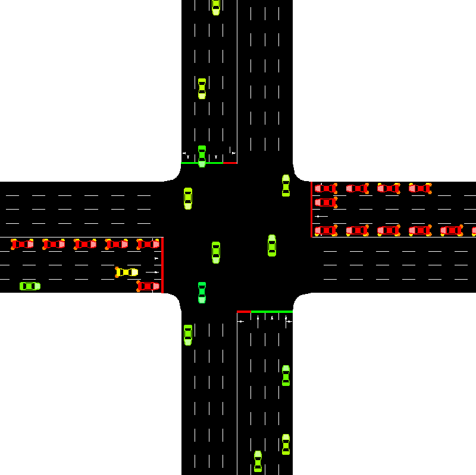}
  \caption{Simulated isolated intersection in SUMO}
  \label{fig:intersection}
\end{figure}

Experiments were conducted on a simulated four-leg intersection in SUMO (Figure~\ref{fig:intersection}). Each approach is 400\,m long with four lanes: one left-turn, two through, and one shared through–right lane. The yellow interval is 3\,s and the all-red interval is 1\,s. Minimum greens are 20\,s for through and 15\,s for left turns.

The goal is to evaluate whether the proposed method improves the algorithm's performance under heavy demand. The intersection operates with four fixed phases: east-west (EW) through, EW left-turn, north-south (NS) through, and NS left-turn phases. Three scenarios (S1–S3) represent distinct demand patterns with overall heavy traffic conditions. Scenario~S1 is balanced; S2 emphasizes NS flows; S3 emphasizes EW through and NS left. Demands and Critical Ratio Sum (CRS) values are listed in Table~\ref{tab:traffic_scenarios}.

\begin{table}
    \centering
    \caption{Traffic demand scenarios and Critical Ratio Sum (CRS). All values are in \textit{veh/h}, formatted as \textit{through/left/right}.}
    \small
    \begin{tabular}{cccc}
        \toprule
        \textbf{Scenario} & \textbf{S1} & \textbf{S2} & \textbf{S3} \\
        \midrule
        North (N) & 1550/210/30 & 2600/230/30 & 2500/80/60 \\
        South (S) & 1450/180/50 & 2450/250/50 & 2400/90/70 \\
        East (E)  & 1450/200/40 & 600/80/40   & 400/330/150 \\
        West (W)  & 1400/180/40 & 650/90/40   & 450/340/120 \\
        CRS\textsuperscript{a} & 0.868 & 0.865 & 0.872 \\
        \bottomrule
    \end{tabular}%
    \label{tab:traffic_scenarios}
    \vspace{0.2em}
    \parbox{\linewidth}{\footnotesize \textsuperscript{a} CRS, based on Webster, reflects demand--capacity ratio; values above 0.85 indicate congestion.}
\end{table}

Webster's Python implementation serves as the initial program. The \textit{LLMs Ensemble} samples from DeepSeek-v3~\citep{liu2024deepseek} (40\%), DeepSeek-r1~\citep{guo2025deepseek} (10\%), OpenAI-o4-mini-high (40\%), and OpenAI-o3 (10\%), with temperature 0.6.

\subsection{Evaluation Metrics}
\label{sec:evaluation_metrics}

For each scenario, hourly demands were randomly generated in three sets (each 1800\,s), and average results were used. Overall performance is the mean across scenarios. Two primary measures were considered: average delay \(\bar{d}\) (s/veh) and average stops \(\bar{s}\) (stops/veh).  

For optimization, they were normalized into scores:  
\begin{equation}
S_d = \frac{1}{1 + \bar{d}/100}, \quad
S_s = \frac{1}{1 + \bar{s}},
\end{equation}
and combined as  
\begin{equation}
S_c = 0.5 \times S_d + 0.5 \times S_s.
\end{equation}
Higher \(S_c\) indicates better performance.

\subsection{Experiment Results}
\label{sec:result}

It is important to note that the objective of this study is not to benchmark against all existing fixed-time optimization methods, but to evaluate whether LLM-guided program evolution can improve upon a classical and interpretable baseline. The use of Webster's method as the initial program provides a transparent and widely recognized reference point for assessing improvements.

\begin{figure}
  \centering
  \includegraphics[width=0.55\textwidth]{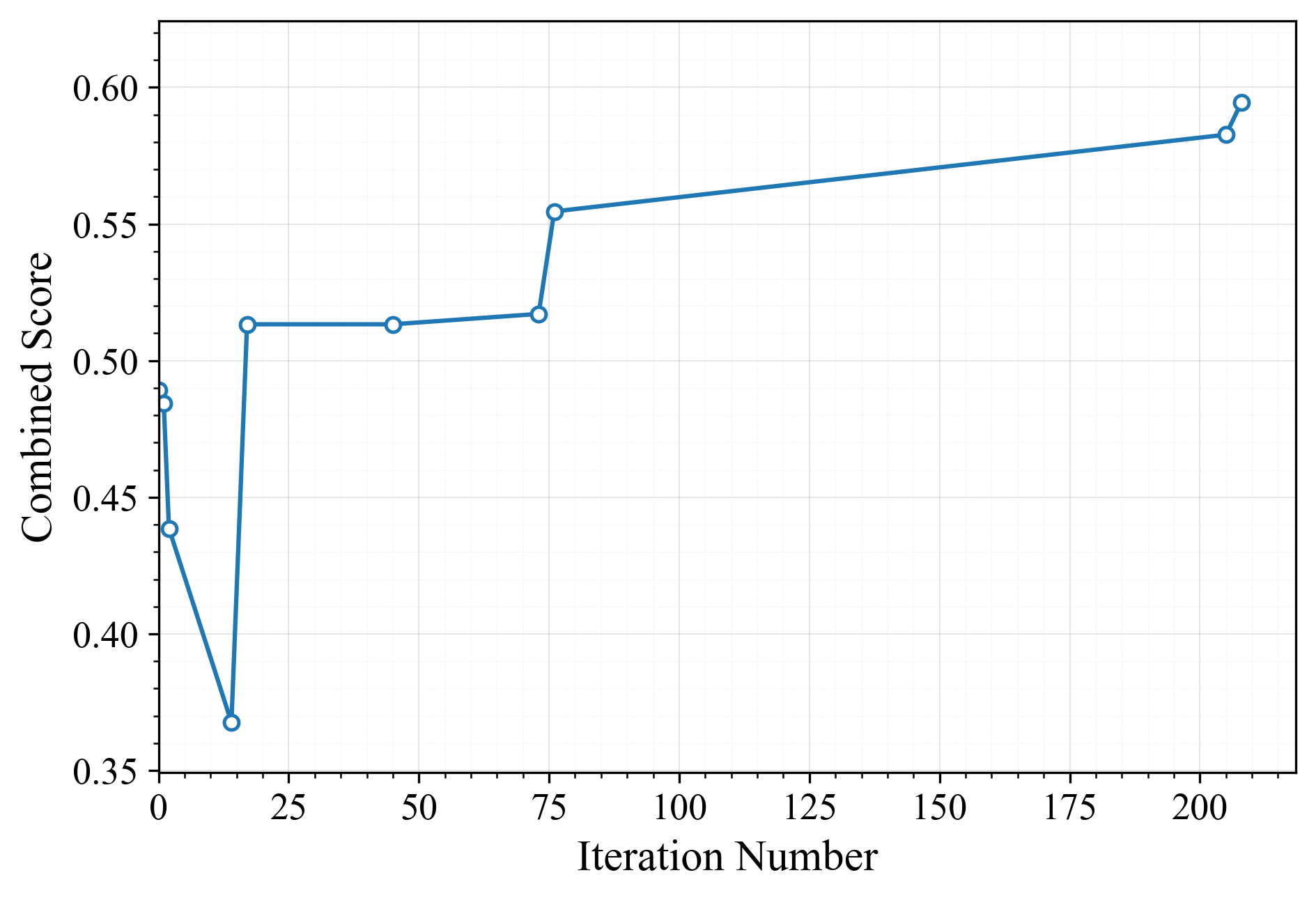}
  \caption{Evolution path of the best discovered program}
  \label{fig:evolution}
\end{figure}

Figure~\ref{fig:evolution} shows the evolution path of the best discovered program after 300 iterations, along with its performance. The performance generally shows an upward trend with occasional plateaus or even declines during some iterations. The initial performance drop reflects exploration in early iterations, where diverse but suboptimal candidate programs are generated before convergence toward higher-performing solutions. Overall, as shown in Table~\ref{tab:performance_comparison}, the combined score improved from 0.4893 (for the initial program) to 0.5946, which represents a 21.5\% improvement. The optimized program outperforms the initial program with a 20.1\% reduction in average delay and a 47.1\% reduction in the number of stops, demonstrating the effectiveness of the evolutionary approach.

\begin{table}[h!]
    \centering
    \caption{Performance Comparison Between the Initial Program and the Best Discovered Program After 300 Iterations}
    \small
    \begin{tabular}{cccc}
        \toprule
        \textbf{Program} & \(\bar{d}\) & \(\bar{s}\) & \textbf{\(S_c\)} \\
        \midrule
        Initial Program & 91.79 & 1.19 & 0.4893 \\
        Discovered Program & 73.31 & 0.63 & 0.5946 \\
        \midrule
        \textbf{Improvement (\%)} & \textbf{-20.1\%} & \textbf{-47.1\%} & \textbf{+21.5\%} \\
        \bottomrule
    \end{tabular}%
    
    \label{tab:performance_comparison}
\end{table}

Figure~\ref{fig:initial_program} and Figure~\ref{fig:optimized_program} present the complete Python code of the initial program and the best discovered program. The discovered program introduces several key modifications (highlighted by blue dashed lines in Figure~\ref{fig:optimized_program}), with the underlying reasoning provided by the LLMs:

\begin{itemize}
    \item \textbf{Cycle Length Bound (CLB)}. The maximum cycle length was increased from 130~s to 240~s. As shown in Figure~\ref{fig:CLB}, the LLM argued that a longer cycle allows more effective green time in high-demand scenarios, thereby reducing delay and stops.

    \item \textbf{Right-Turn Inclusion (RTI)}. Right-turn flows were added to the through demand calculation. Figure~\ref{fig:RTI_and_PAR} illustrates how the LLM observed that right-turn vehicles were omitted in the initial program. Given the presence of through–right shared lanes, it is suggested to include right-turn flows to better balance capacity.

    \item \textbf{Shared Lane Factor (SLF)}. The contribution of the shared through–right lane was reduced from 0.9 to 0.5. As shown in Figure~\ref{fig:SLF}, the LLM reasoned that a factor of 0.9 overestimated lane capacity, which underestimated the required green time for through movements. Reducing the factor to 0.5 raised their priority, allocating more green to critical through directions.

    \item \textbf{Minimum-Green Feasibility (MGF)}. The cycle length was adjusted when the effective green was insufficient to satisfy mandatory minimums. The LLM noted that the initial thresholds (50, 90) were too short to cover the required minimums (20~s per through phase and 15~s per left-turn, plus 16~s for intergreens, totaling 86~s). It accordingly revised the range to (90, 140) to ensure feasibility.

    \item \textbf{Post-Allocation Rescaling (PAR)}. Green times were proportionally rescaled to match the effective green budget. As shown in Figure~\ref{fig:RTI_and_PAR}, this avoided both unused “dead” green seconds, which increase delay, and overruns, which reduce capacity for other phases.
\end{itemize}

\paragraph{Ablation from the discovered program}
Table~\ref{tab:ablation} shows the ablation results when removing individual modifications from the discovered program. Among all components, \textbf{CLB} is the dominant contributor: its removal decreases \(S_c\) from \(0.5946\) to \(0.4812\) (\(-19.07\%\)). The benefit of CLB stems from longer cycles diluting the fixed 16\,s lost time, thereby increasing the share of effective green under heavy demand. Crucially, the improvement does not arise from arbitrarily lengthening cycles but from avoiding performance degradation due to overly restrictive bounds. \textbf{RTI} and \textbf{SLF} provide secondary yet important gains: removing either reduces \(S_c\) by \(3.48\%\) and \(2.98\%\), respectively, reflecting their role in balancing green allocation by accounting for right-turn demand and correcting shared-lane capacity. \textbf{PAR} offers a modest but stable gain (\(-0.81\%\) when removed), largely due to rounding effects in high-demand scenarios; it would be more impactful under heterogeneous traffic when some phases approach minimum-green thresholds. Finally, \textbf{MGF} has no effect here since all tested cycles are sufficiently long, though it is expected to safeguard feasibility under lighter traffic.

\paragraph{Incremental builds from the initial program}
To further analyze the most effective strategies, Table~\ref{tab:incremental} presents an incremental analysis starting from the initial program. Again, \textbf{CLB} dominates: adding CLB alone improves \(S_c\) by \(15.63\%\). Adding \textbf{SLF} yields a minor gain (\(+1.51\%\)), while adding \textbf{RTI} alone surprisingly degrades performance (\(-2.27\%\)), as incorporating extra demand without cycle or capacity adjustment distorts green allocation. Notably, combining modifications yields synergistic rather than additive effects. For instance, \texttt{CLB+RTI} and \texttt{CLB+SLF} improve \(S_c\) by \(+17.90\%\) and \(+17.29\%\), respectively, exceeding the sum of individual contributions. This suggests a coupling between demand modeling and cycle flexibility: RTI or SLF only becomes effective when longer cycles allow sufficient redistribution of green. These findings highlight two insights: (i) some modifications are detrimental in isolation, and (ii) evolutionary search is crucial for identifying synergistic combinations that outperform individual strategies. Beyond performance, this also illustrates how a coding-agent framework can uncover non-intuitive design patterns that may be difficult for human engineers to identify through manual, additive reasoning.

\begin{table}[h!]
    \centering
    \caption{Ablation study of the optimized program by removing individual modifications. (``w/o'' denotes removal of a component).}
    \small
    \begin{tabular}{lcccc}
        \toprule
        \textbf{Variant} & \(\bar{d}\) & \(\bar{s}\) & \(S_c\) & Change \\
        \midrule
        Optimized Program & 73.31 & 0.63 & 0.5946 & 0.00\% \\
        \; w/o CLB & 89.32 & 1.30 & 0.4812 & -19.07\% \\
        \; w/o RTI & 81.00 & 0.68 & 0.5739 & -3.48\% \\
        \; w/o SLF & 76.46 & 0.70 & 0.5769 & -2.98\% \\
        \; w/o MGF & 73.31 & 0.63 & 0.5946 & 0.00\% \\
        \; w/o PAR & 73.64 & 0.66 & 0.5898 & -0.81\% \\
        \bottomrule
    \end{tabular}
    
    \label{tab:ablation}
\end{table}

\begin{table}[h!]
    \centering
    \caption{Incremental analysis by progressively adding modifications to the initial program. (``+'' denotes addition of a component)}
    \small
    \begin{tabular}{lcccc}
        \toprule
        \textbf{Variant} & \(\bar{d}\) & \(\bar{s}\) & \(S_c\) & Change \\
        \midrule
        Initial Program & 91.79 & 1.19 & 0.4893 & 0.00\% \\
        \; + CLB & 84.01 & 0.70 & 0.5658 & +15.63\% \\
        \; + RTI & 90.81 & 1.31 & 0.4782 & -2.27\% \\
        \; + SLF & 87.27 & 1.18 & 0.4967 & +1.51\% \\
        \; + CLB + RTI & 76.46 & 0.70 & 0.5769 & +17.90\% \\
        \; + RTI + SLF & 87.56 & 1.20 & 0.4935 & +0.86\% \\
        \; + CLB + SLF & 81.00 & 0.68 & 0.5739 & +17.29\% \\
        \bottomrule
    \end{tabular}
    
    \label{tab:incremental}
\end{table}

\section{Discussion}
\label{sec:discussion}

\subsection{Types of LLM-Generated Modifications}
Across iterations, most LLM-generated modifications fall into three categories: \emph{heuristic rules}, \emph{traffic-engineering principles}, and \emph{Hyper-parameter adjustments}. The first introduces conditional rules (e.g., \texttt{if--else} branches) to adapt logic under different demand conditions. The second leverages domain knowledge, such as refining critical movement analysis or adjusting green allocation to reflect imbalances. The third modifies parameters such as cycle length, saturation flow, or shared-lane factors; unlike conventional tuning, these changes are typically justified with reasoning about expected effects. Importantly, not all useful modifications remain in the best discovered program. As ablation and incremental analyses show, some are only beneficial in combination, while others degrade performance in isolation. Even discarded strategies (e.g., Figure~\ref{fig:discarded_modifications}) can be valuable, as refinements in saturation flow or demand-scaled allocation highlight potential directions for practical design.

\subsection{Exploration and Diversity}
Increasing iterations and adding more models to the \textit{LLMs Ensemble} can further improve outcomes. The extended run in Figure~\ref{fig:evolution600} achieved slightly higher scores, reflecting the benefit of broader exploration. Different LLMs often suggest distinct modifications, enhancing diversity and the chance of synergy. However, more modifications also make code harder to interpret and maintain, which may hinder deployment. To balance this, we focused on a relatively interpretable discovered program, while recognizing that multiple runs may produce different yet comparable solutions. Under a fixed prompt template, the same LLM (e.g., DeepSeek-r1) tends to generate similar families of modifications even with relative large temperatures, revealing stable biases; this consistency implies that greater diversity must come from varying prompts or combining multiple LLMs.

\subsection{Interpretability and Practical Value}
Although the internal reasoning of LLMs is black-box, the generated modifications are accompanied by natural language explanations that align with traffic engineering logic. This provides interpretability absent in conventional black-box optimization. Moreover, since outputs are executable Python code, practitioners can directly inspect and adapt modifications. This combination of conceptual reasoning and explicit code offers practical value, enabling engineers to retain or refine the most useful strategies. In practice, such human-in-the-loop supervision may prove as valuable as raw performance gains, bridging AI-driven synthesis with domain expertise in signal control.

\section{Conclusion}
\label{sec:conclusion}

This paper presented \textbf{EvolveSignal}, an LLM-powered framework for the automated discovery of interpretable heuristic strategies for fixed-time traffic signal control. By formulating fixed-time optimization as a program synthesis problem, the framework moves beyond conventional parameter tuning and enables the exploration of code-level modifications expressed as executable programs. Experiments on a congested intersection showed that the discovered strategy outperformed Webster's method, reducing average delay by 20.1\% and stops by 47.1\%.

Ablation and incremental analyses indicate that not all modifications are effective in isolation, but certain combinations can yield synergistic improvements. Even discarded modifications provide interpretable insights, illustrating how LLM-driven exploration can support the development of practical traffic engineering heuristics. In addition, the framework produces human-readable code accompanied by natural language reasoning, allowing practitioners to inspect, refine, and adapt the results.

This study represents an initial step toward LLM-driven heuristic discovery in traffic signal control. The current evaluation is limited to a single intersection and a small set of demand scenarios, which may not fully capture the diversity of real-world conditions. Future work will extend the framework to more complex network-level control, broader traffic patterns, and additional evaluation criteria such as fairness, safety, and operational constraints. Comparisons with a wider range of baseline methods and alternative program search strategies (e.g., random or rule-based mutations) will further clarify the contribution of LLM-guided synthesis.

\section*{Acknowledgments}

The work was supported by start-up funds (No. MSRI8001004 and No. MSRI9002005) at Monash University, as well as by the TRENoP and Digital Futures research centers at KTH Royal Institute of Technology, Sweden.

\section*{AUTHOR CONTRIBUTIONS}
The authors confirm contribution to the paper as follows: study conception and design: Z Ma, L Wang, H Wang, Y Wang, P Duan, N Zheng; methodology: P Duan, L Wang, Z Ma, H Wang, J Xu; data collection: L Wang, Y Wang, J Xu; analysis and interpretation of results: L Wang, P Duan, Z Ma, H Wang, Y Wang; draft manuscript preparation: L Wang, P Duan, Z Ma, H Wang, Y Wang, J Xu; manuscript revision: L Wang, P Duan, Z Ma, H Wang, Y Wang, J Xu, N Zheng. All authors reviewed the results and approved the final version of the manuscript.



\bibliographystyle{elsarticle-harv} 
\bibliography{cas-refs}

\begin{thebibliography}{36}
\expandafter\ifx\csname natexlab\endcsname\relax\def\natexlab#1{#1}\fi
\providecommand{\url}[1]{\texttt{#1}}
\providecommand{\href}[2]{#2}
\providecommand{\path}[1]{#1}
\providecommand{\DOIprefix}{doi:}
\providecommand{\ArXivprefix}{arXiv:}
\providecommand{\URLprefix}{URL: }
\providecommand{\Pubmedprefix}{pmid:}
\providecommand{\doi}[1]{\href{http://dx.doi.org/#1}{\path{#1}}}
\providecommand{\Pubmed}[1]{\href{pmid:#1}{\path{#1}}}
\providecommand{\bibinfo}[2]{#2}
\ifx\xfnm\relax \def\xfnm[#1]{\unskip,\space#1}\fi
\bibitem[{Chu et~al.(2019)Chu, Wang, Codec{\`a} and Li}]{chu2019multi}
\bibinfo{author}{Chu, T.}, \bibinfo{author}{Wang, J.}, \bibinfo{author}{Codec{\`a}, L.}, \bibinfo{author}{Li, Z.}, \bibinfo{year}{2019}.
\newblock \bibinfo{title}{Multi-agent deep reinforcement learning for large-scale traffic signal control}.
\newblock \bibinfo{journal}{IEEE transactions on intelligent transportation systems} \bibinfo{volume}{21}, \bibinfo{pages}{1086--1095}.
\bibitem[{Fellendorf(1994)}]{fellendorf1994vissim}
\bibinfo{author}{Fellendorf, M.}, \bibinfo{year}{1994}.
\newblock \bibinfo{title}{Vissim: A microscopic simulation tool to evaluate actuated signal control including bus priority}, in: \bibinfo{booktitle}{64th Institute of transportation engineers annual meeting}, \bibinfo{organization}{Springer Berlin/Heidelberg, Germany}. pp. \bibinfo{pages}{1--9}.
\bibitem[{Gartner et~al.(1991)Gartner, Assman, Lasaga and Hou}]{gartner1991multi}
\bibinfo{author}{Gartner, N.H.}, \bibinfo{author}{Assman, S.F.}, \bibinfo{author}{Lasaga, F.}, \bibinfo{author}{Hou, D.L.}, \bibinfo{year}{1991}.
\newblock \bibinfo{title}{A multi-band approach to arterial traffic signal optimization}.
\newblock \bibinfo{journal}{Transportation Research Part B: Methodological} \bibinfo{volume}{25}, \bibinfo{pages}{55--74}.
\bibitem[{Gong et~al.(2020)Gong, Abdel-Aty, Yuan and Cai}]{gong2020multi}
\bibinfo{author}{Gong, Y.}, \bibinfo{author}{Abdel-Aty, M.}, \bibinfo{author}{Yuan, J.}, \bibinfo{author}{Cai, Q.}, \bibinfo{year}{2020}.
\newblock \bibinfo{title}{Multi-objective reinforcement learning approach for improving safety at intersections with adaptive traffic signal control}.
\newblock \bibinfo{journal}{Accident Analysis \& Prevention} \bibinfo{volume}{144}, \bibinfo{pages}{105655}.
\bibitem[{Guo et~al.(2025)Guo, Yang, Zhang, Song, Zhang, Xu, Zhu, Ma, Wang, Bi et~al.}]{guo2025deepseek}
\bibinfo{author}{Guo, D.}, \bibinfo{author}{Yang, D.}, \bibinfo{author}{Zhang, H.}, \bibinfo{author}{Song, J.}, \bibinfo{author}{Zhang, R.}, \bibinfo{author}{Xu, R.}, \bibinfo{author}{Zhu, Q.}, \bibinfo{author}{Ma, S.}, \bibinfo{author}{Wang, P.}, \bibinfo{author}{Bi, X.}, et~al., \bibinfo{year}{2025}.
\newblock \bibinfo{title}{Deepseek-r1: Incentivizing reasoning capability in llms via reinforcement learning}.
\newblock \bibinfo{journal}{arXiv preprint arXiv:2501.12948} .
\bibitem[{Guo et~al.(2024)Guo, Zhang, Jiang, Peng, Zhu and Yang}]{guo2024towards}
\bibinfo{author}{Guo, X.}, \bibinfo{author}{Zhang, Q.}, \bibinfo{author}{Jiang, J.}, \bibinfo{author}{Peng, M.}, \bibinfo{author}{Zhu, M.}, \bibinfo{author}{Yang, H.F.}, \bibinfo{year}{2024}.
\newblock \bibinfo{title}{Towards explainable traffic flow prediction with large language models}.
\newblock \bibinfo{journal}{Communications in Transportation Research} \bibinfo{volume}{4}, \bibinfo{pages}{100150}.
\bibitem[{Hunt et~al.(1982)Hunt, Robertson, Bretherton and Royle}]{hunt1982scoot}
\bibinfo{author}{Hunt, P.}, \bibinfo{author}{Robertson, D.}, \bibinfo{author}{Bretherton, R.}, \bibinfo{author}{Royle, M.C.}, \bibinfo{year}{1982}.
\newblock \bibinfo{title}{The scoot on-line traffic signal optimisation technique}.
\newblock \bibinfo{journal}{Traffic Engineering \& Control} \bibinfo{volume}{23}.
\bibitem[{Jin et~al.(2026)Jin, Wu, Zhang, Zhang, Yin and Dong}]{jin2026hybrid}
\bibinfo{author}{Jin, J.}, \bibinfo{author}{Wu, S.}, \bibinfo{author}{Zhang, P.}, \bibinfo{author}{Zhang, X.}, \bibinfo{author}{Yin, C.}, \bibinfo{author}{Dong, C.}, \bibinfo{year}{2026}.
\newblock \bibinfo{title}{Hybrid action space approach to traffic signal optimization using deep reinforcement learning}.
\newblock \bibinfo{journal}{Computers \& Operations Research} , \bibinfo{pages}{107391}.
\bibitem[{Koch et~al.(2023)Koch, Brinkmann, Wegener, Badalian and Andert}]{koch2023adaptive}
\bibinfo{author}{Koch, L.}, \bibinfo{author}{Brinkmann, T.}, \bibinfo{author}{Wegener, M.}, \bibinfo{author}{Badalian, K.}, \bibinfo{author}{Andert, J.}, \bibinfo{year}{2023}.
\newblock \bibinfo{title}{Adaptive traffic light control with deep reinforcement learning: An evaluation of traffic flow and energy consumption}.
\newblock \bibinfo{journal}{IEEE transactions on intelligent transportation systems} \bibinfo{volume}{24}, \bibinfo{pages}{15066--15076}.
\bibitem[{Kuwahara et~al.(2025)Kuwahara, Hashimoto, Tanabe and Yoshioka}]{kuwahara2025optimum}
\bibinfo{author}{Kuwahara, M.}, \bibinfo{author}{Hashimoto, S.}, \bibinfo{author}{Tanabe, J.}, \bibinfo{author}{Yoshioka, K.}, \bibinfo{year}{2025}.
\newblock \bibinfo{title}{Optimum traffic control by decentralized reinforcement learning utilizing kinematic wave propagation-applications to traffic signal control}.
\newblock \bibinfo{journal}{Artificial Intelligence for Transportation} \bibinfo{volume}{3}, \bibinfo{pages}{100036}.
\bibitem[{Lai et~al.(2025)Lai, Xu, Zhang, Liu and Xiong}]{lai2025llmlight}
\bibinfo{author}{Lai, S.}, \bibinfo{author}{Xu, Z.}, \bibinfo{author}{Zhang, W.}, \bibinfo{author}{Liu, H.}, \bibinfo{author}{Xiong, H.}, \bibinfo{year}{2025}.
\newblock \bibinfo{title}{Llmlight: Large language models as traffic signal control agents}, in: \bibinfo{booktitle}{Proceedings of the 31st ACM SIGKDD Conference on Knowledge Discovery and Data Mining V. 1}, pp. \bibinfo{pages}{2335--2346}.
\bibitem[{Ling et~al.(2026)Ling, Qin and Ma}]{ling2026review}
\bibinfo{author}{Ling, Y.}, \bibinfo{author}{Qin, Z.}, \bibinfo{author}{Ma, Z.}, \bibinfo{year}{2026}.
\newblock \bibinfo{title}{A review of knowledge graph construction using large language models in transportation: Problems, methods, and challenges}.
\newblock \bibinfo{journal}{Transportation Research Part C: Emerging Technologies} \bibinfo{volume}{183}, \bibinfo{pages}{105428}.
\bibitem[{Liu et~al.(2024)Liu, Feng, Xue, Wang, Wu, Lu, Zhao, Deng, Zhang, Ruan et~al.}]{liu2024deepseek}
\bibinfo{author}{Liu, A.}, \bibinfo{author}{Feng, B.}, \bibinfo{author}{Xue, B.}, \bibinfo{author}{Wang, B.}, \bibinfo{author}{Wu, B.}, \bibinfo{author}{Lu, C.}, \bibinfo{author}{Zhao, C.}, \bibinfo{author}{Deng, C.}, \bibinfo{author}{Zhang, C.}, \bibinfo{author}{Ruan, C.}, et~al., \bibinfo{year}{2024}.
\newblock \bibinfo{title}{Deepseek-v3 technical report}.
\newblock \bibinfo{journal}{arXiv preprint arXiv:2412.19437} .
\bibitem[{Liu et~al.(2025)Liu, Yang and Yin}]{liu2025toward}
\bibinfo{author}{Liu, T.}, \bibinfo{author}{Yang, J.}, \bibinfo{author}{Yin, Y.}, \bibinfo{year}{2025}.
\newblock \bibinfo{title}{Toward llm-agent-based modeling of transportation systems: A conceptual framework}.
\newblock \bibinfo{journal}{Artificial Intelligence for Transportation} \bibinfo{volume}{1}, \bibinfo{pages}{100001}.
\bibitem[{Lopez et~al.(2018)Lopez, Behrisch, Bieker-Walz, Erdmann, Fl{\"o}tter{\"o}d, Hilbrich, L{\"u}cken, Rummel, Wagner and Wie{\ss}ner}]{lopez2018microscopic}
\bibinfo{author}{Lopez, P.A.}, \bibinfo{author}{Behrisch, M.}, \bibinfo{author}{Bieker-Walz, L.}, \bibinfo{author}{Erdmann, J.}, \bibinfo{author}{Fl{\"o}tter{\"o}d, Y.P.}, \bibinfo{author}{Hilbrich, R.}, \bibinfo{author}{L{\"u}cken, L.}, \bibinfo{author}{Rummel, J.}, \bibinfo{author}{Wagner, P.}, \bibinfo{author}{Wie{\ss}ner, E.}, \bibinfo{year}{2018}.
\newblock \bibinfo{title}{Microscopic traffic simulation using sumo}, in: \bibinfo{booktitle}{2018 21st international conference on intelligent transportation systems (ITSC)}, \bibinfo{organization}{Ieee}. pp. \bibinfo{pages}{2575--2582}.
\bibitem[{Ma et~al.(2026)Ma, Wang, Qin and Ling}]{ma2026large}
\bibinfo{author}{Ma, Z.}, \bibinfo{author}{Wang, L.}, \bibinfo{author}{Qin, Z.}, \bibinfo{author}{Ling, Y.}, \bibinfo{year}{2026}.
\newblock \bibinfo{title}{Large language models for urban transportation}, in: \bibinfo{booktitle}{Mobility Patterns, Big Data and Transport Analytics}. \bibinfo{publisher}{Elsevier}, pp. \bibinfo{pages}{287--318}.
\bibitem[{Mouret and Clune(2015)}]{mouret2015illuminating}
\bibinfo{author}{Mouret, J.B.}, \bibinfo{author}{Clune, J.}, \bibinfo{year}{2015}.
\newblock \bibinfo{title}{Illuminating search spaces by mapping elites}.
\newblock \bibinfo{journal}{arXiv preprint arXiv:1504.04909} .
\bibitem[{Movahedi and Choi(2024)}]{movahedi2024crossroads}
\bibinfo{author}{Movahedi, M.}, \bibinfo{author}{Choi, J.}, \bibinfo{year}{2024}.
\newblock \bibinfo{title}{The crossroads of llm and traffic control: A study on large language models in adaptive traffic signal control}.
\newblock \bibinfo{journal}{IEEE Transactions on Intelligent Transportation Systems} .
\bibitem[{Nie et~al.(2025)Nie, Sun and Ma}]{nie2025exploring}
\bibinfo{author}{Nie, T.}, \bibinfo{author}{Sun, J.}, \bibinfo{author}{Ma, W.}, \bibinfo{year}{2025}.
\newblock \bibinfo{title}{Exploring the roles of large language models in reshaping transportation systems: A survey, framework, and roadmap}.
\newblock \bibinfo{journal}{Artificial Intelligence for Transportation} \bibinfo{volume}{1}, \bibinfo{pages}{100003}.
\bibitem[{Novikov et~al.(2025)Novikov, V{\~u}, Eisenberger, Dupont, Huang, Wagner, Shirobokov, Kozlovskii, Ruiz, Mehrabian et~al.}]{novikov2025alphaevolve}
\bibinfo{author}{Novikov, A.}, \bibinfo{author}{V{\~u}, N.}, \bibinfo{author}{Eisenberger, M.}, \bibinfo{author}{Dupont, E.}, \bibinfo{author}{Huang, P.S.}, \bibinfo{author}{Wagner, A.Z.}, \bibinfo{author}{Shirobokov, S.}, \bibinfo{author}{Kozlovskii, B.}, \bibinfo{author}{Ruiz, F.J.}, \bibinfo{author}{Mehrabian, A.}, et~al., \bibinfo{year}{2025}.
\newblock \bibinfo{title}{Alphaevolve: A coding agent for scientific and algorithmic discovery}.
\newblock \bibinfo{journal}{arXiv preprint arXiv:2506.13131} .
\bibitem[{Pang et~al.(2026)Pang, Patel, Liu, Wang, Lin, He and Li}]{pang2026large}
\bibinfo{author}{Pang, S.}, \bibinfo{author}{Patel, D.}, \bibinfo{author}{Liu, Y.}, \bibinfo{author}{Wang, S.}, \bibinfo{author}{Lin, C.}, \bibinfo{author}{He, S.}, \bibinfo{author}{Li, T.}, \bibinfo{year}{2026}.
\newblock \bibinfo{title}{Large language models for car following in automated driving: Opportunities and challenges}.
\newblock \bibinfo{journal}{Artificial Intelligence for Transportation} \bibinfo{volume}{6}, \bibinfo{pages}{100047}.
\bibitem[{Peng et~al.(2025)Peng, Chen, Gao, Wang and Zhang}]{peng2025combat}
\bibinfo{author}{Peng, X.}, \bibinfo{author}{Chen, S.}, \bibinfo{author}{Gao, H.}, \bibinfo{author}{Wang, H.}, \bibinfo{author}{Zhang, H.M.}, \bibinfo{year}{2025}.
\newblock \bibinfo{title}{Combat urban congestion via collaboration: Heterogeneous gnn-based marl for coordinated platooning and traffic signal control}.
\newblock \bibinfo{journal}{IEEE Transactions on Intelligent Transportation Systems} .
\bibitem[{Qin et~al.(2025a)Qin, Wang, Pereira and Ma}]{qin2025foundational}
\bibinfo{author}{Qin, Z.}, \bibinfo{author}{Wang, L.}, \bibinfo{author}{Pereira, F.C.}, \bibinfo{author}{Ma, Z.}, \bibinfo{year}{2025}a.
\newblock \bibinfo{title}{A foundational individual mobility prediction model based on open-source large language models}.
\newblock \bibinfo{journal}{arXiv preprint arXiv:2503.16553} .
\bibitem[{Qin et~al.(2025b)Qin, Zhang, Wang and Ma}]{qin2025lingotrip}
\bibinfo{author}{Qin, Z.}, \bibinfo{author}{Zhang, P.}, \bibinfo{author}{Wang, L.}, \bibinfo{author}{Ma, Z.}, \bibinfo{year}{2025}b.
\newblock \bibinfo{title}{Lingotrip: Spatiotemporal context prompt driven large language model for individual trip prediction}.
\newblock \bibinfo{journal}{Journal of Public Transportation} \bibinfo{volume}{27}, \bibinfo{pages}{100117}.
\bibitem[{Ren et~al.(2025)Ren, Chang, Cui, Chang, Yu, Li and Wang}]{ren2025cooperative}
\bibinfo{author}{Ren, Y.}, \bibinfo{author}{Chang, Y.}, \bibinfo{author}{Cui, Z.}, \bibinfo{author}{Chang, X.}, \bibinfo{author}{Yu, H.}, \bibinfo{author}{Li, X.}, \bibinfo{author}{Wang, Y.}, \bibinfo{year}{2025}.
\newblock \bibinfo{title}{Is cooperative always better? multi-agent reinforcement learning with explicit neighborhood backtracking for network-wide traffic signal control}.
\newblock \bibinfo{journal}{Transportation Research Part C: Emerging Technologies} \bibinfo{volume}{179}, \bibinfo{pages}{105265}.
\bibitem[{Sharma(2025)}]{openevolve}
\bibinfo{author}{Sharma, A.}, \bibinfo{year}{2025}.
\newblock \bibinfo{title}{Openevolve: an open-source evolutionary coding agent}.
\newblock \URLprefix \url{https://github.com/codelion/openevolve}.
\bibitem[{Wang and Peng(2022)}]{wang2022coordinated}
\bibinfo{author}{Wang, H.}, \bibinfo{author}{Peng, X.}, \bibinfo{year}{2022}.
\newblock \bibinfo{title}{Coordinated control model for oversaturated arterial intersections}.
\newblock \bibinfo{journal}{IEEE Transactions on Intelligent Transportation Systems} \bibinfo{volume}{23}, \bibinfo{pages}{24157--24175}.
\bibitem[{Wang et~al.(2025a)Wang, Duan, He, Lyu, Chen, Zheng, Yao and Ma}]{wang2025agentic}
\bibinfo{author}{Wang, L.}, \bibinfo{author}{Duan, P.}, \bibinfo{author}{He, Z.}, \bibinfo{author}{Lyu, C.}, \bibinfo{author}{Chen, X.}, \bibinfo{author}{Zheng, N.}, \bibinfo{author}{Yao, L.}, \bibinfo{author}{Ma, Z.}, \bibinfo{year}{2025}a.
\newblock \bibinfo{title}{Agentic large language models for day-to-day route choices}.
\newblock \bibinfo{journal}{Transportation Research Part C: Emerging Technologies} \bibinfo{volume}{180}, \bibinfo{pages}{105307}.
\bibitem[{Wang et~al.(2023)Wang, Ma, Dong and Wang}]{wang2023human}
\bibinfo{author}{Wang, L.}, \bibinfo{author}{Ma, Z.}, \bibinfo{author}{Dong, C.}, \bibinfo{author}{Wang, H.}, \bibinfo{year}{2023}.
\newblock \bibinfo{title}{Human-centric multimodal deep (hmd) traffic signal control}.
\newblock \bibinfo{journal}{IET Intelligent Transport Systems} \bibinfo{volume}{17}, \bibinfo{pages}{744--753}.
\bibitem[{Wang et~al.(2018)Wang, Yang, Liang and Liu}]{wang2018review}
\bibinfo{author}{Wang, Y.}, \bibinfo{author}{Yang, X.}, \bibinfo{author}{Liang, H.}, \bibinfo{author}{Liu, Y.}, \bibinfo{year}{2018}.
\newblock \bibinfo{title}{A review of the self-adaptive traffic signal control system based on future traffic environment}.
\newblock \bibinfo{journal}{Journal of Advanced Transportation} \bibinfo{volume}{2018}, \bibinfo{pages}{1096123}.
\bibitem[{Wang et~al.(2025b)Wang, Zhou, Huang, Zhuo, Yi and Ma}]{wang2025chat2spat}
\bibinfo{author}{Wang, Y.}, \bibinfo{author}{Zhou, M.}, \bibinfo{author}{Huang, G.}, \bibinfo{author}{Zhuo, R.}, \bibinfo{author}{Yi, C.}, \bibinfo{author}{Ma, Z.}, \bibinfo{year}{2025}b.
\newblock \bibinfo{title}{Chat2spat: A large language model based tool for automating traffic signal control plan management}.
\newblock \bibinfo{journal}{arXiv preprint arXiv:2507.05283} .
\bibitem[{Webster(1958)}]{webster1958traffic}
\bibinfo{author}{Webster, F.V.}, \bibinfo{year}{1958}.
\newblock \bibinfo{title}{Traffic signal settings}.
\newblock \bibinfo{type}{Technical Report}.
\bibitem[{Wei et~al.(2019)Wei, Zheng, Gayah and Li}]{wei2019survey}
\bibinfo{author}{Wei, H.}, \bibinfo{author}{Zheng, G.}, \bibinfo{author}{Gayah, V.}, \bibinfo{author}{Li, Z.}, \bibinfo{year}{2019}.
\newblock \bibinfo{title}{A survey on traffic signal control methods}.
\newblock \bibinfo{journal}{arXiv preprint arXiv:1904.08117} .
\bibitem[{Xie et~al.(2025)Xie, Dong and Wang}]{xie2025coordination}
\bibinfo{author}{Xie, N.}, \bibinfo{author}{Dong, C.}, \bibinfo{author}{Wang, H.}, \bibinfo{year}{2025}.
\newblock \bibinfo{title}{Coordination of distributed adaptive signal control and advisory speed optimization based on shockwave theory}.
\newblock \bibinfo{journal}{Computer-Aided Civil and Infrastructure Engineering} \bibinfo{volume}{40}, \bibinfo{pages}{1675--1687}.
\bibitem[{Zhen and Yang(2025)}]{zhen2025crashsage}
\bibinfo{author}{Zhen, H.}, \bibinfo{author}{Yang, J.J.}, \bibinfo{year}{2025}.
\newblock \bibinfo{title}{Crashsage: A large language model-centered framework for contextual and interpretable traffic crash analysis}.
\newblock \bibinfo{journal}{Artificial Intelligence for Transportation} \bibinfo{volume}{3}, \bibinfo{pages}{100030}.
\bibitem[{Zhou et~al.(2024)Zhou, Wang, Liu, Liu, Zhang and Ma}]{zhou2024traffic}
\bibinfo{author}{Zhou, W.}, \bibinfo{author}{Wang, Y.}, \bibinfo{author}{Liu, M.}, \bibinfo{author}{Liu, T.}, \bibinfo{author}{Zhang, P.}, \bibinfo{author}{Ma, Z.}, \bibinfo{year}{2024}.
\newblock \bibinfo{title}{Traffic signal phase and timing estimation using trajectory data from radar vision integrated camera}.
\newblock \bibinfo{journal}{IEEE Transactions on Intelligent Transportation Systems} \bibinfo{volume}{25}, \bibinfo{pages}{18279--18291}.
\newblock \DOIprefix\doi{10.1109/TITS.2024.3440359}.

\end{thebibliography}






\newpage  

\section*{Appendix}

\subsection{Detailed Description of Framework Modules}
\label{appendix:modules}

This appendix provides extended details of the four modules in the proposed EvolveSignal framework (Section~\ref{sec:method}): \textit{Program Database}, \textit{Prompt Sampler}, \textit{LLMs Ensemble}, and \textit{Evaluators Pool}. These were omitted from the main text due to space constraints but are included here for completeness.

\textbf{Program Database.}
The \textit{Program Database} maintains the population of candidate programs during evolution. For each program, it stores performance metrics, evolutionary history, and metadata. To balance exploration and exploitation, the population is organized using the MAP-Elites~\citep{mouret2015illuminating} algorithm, which preserves diversity while promoting high-performing solutions. In each iteration, one parent program and several inspiration programs are sampled for modification.

\textbf{Prompt Sampler.}
The \textit{Prompt Sampler} constructs the prompts provided to the LLMs. It generates a system message (see Figure~\ref{fig:system_message}) that defines the LLM's role and the task it is expected to perform, along with a user message (Figure~\ref{fig:user_message}) that consolidates the parent program, inspiration programs, and their performance metrics. To support different search strategies, a flexible template mechanism allows prompts to specify either local code modifications (analogous to differential evolution) or full rewrites. This ensures that the LLMs receive both structural and performance context, guiding them to produce relevant modifications.

\textbf{LLMs Ensemble.}
The \textit{LLMs Ensemble} integrates multiple LLMs (e.g., DeepSeek-v3~\citep{liu2024deepseek}, DeepSeek-r1~\citep{guo2025deepseek}). In each iteration, one model is sampled to generate suggestions. This diversity increases the likelihood of discovering effective modifications, as different models emphasize different reasoning paths. Figure~\ref{fig:response} shows an example response. Unlike conventional crossover and mutation restricted to parameter values, LLMs can propose structural code-level changes, expanding the search space and enabling novel solutions.

\textbf{Evaluators Pool.}
The \textit{Evaluators Pool} assesses the quality of generated programs using microscopic traffic simulation in SUMO~\citep{lopez2018microscopic}. Each program generates a fixed-time signal control plan (e.g., green allocations per phase) based on the given scenario inputs. This strategy is then executed in simulation, and performance metrics such as average delay and number of stops are recorded. The evaluation results are returned to the \textit{Program Database}, where they influence program retention and guide subsequent modifications.

\subsection{Prompt and Response Examples}
\label{appendix:prompt_and_examples}

\begin{figure}[ht]
  \centering
  \includegraphics[width=0.8\textwidth]{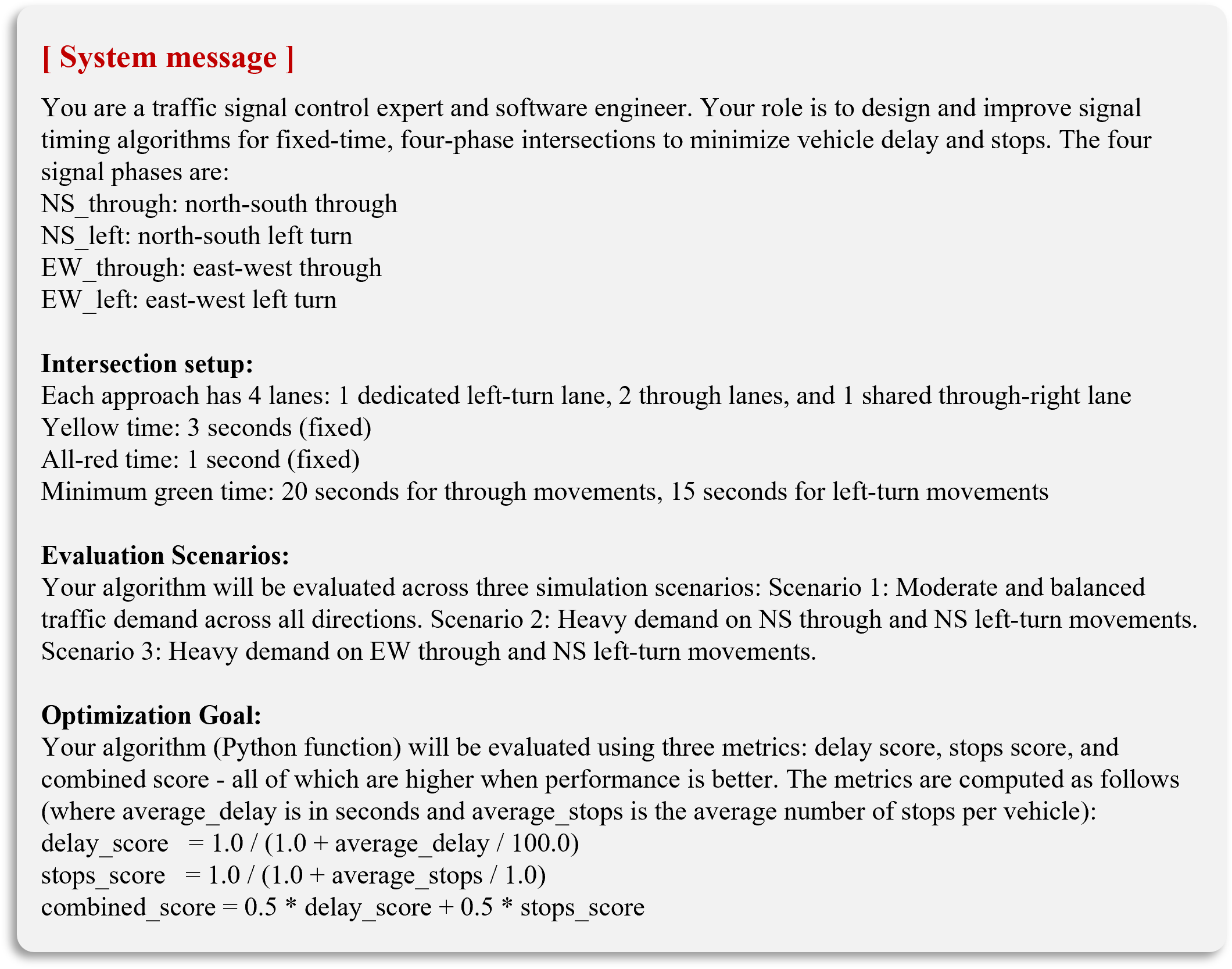}
  \caption{Example system message prompt template}
  \label{fig:system_message}
\end{figure}

\begin{figure}[ht]
  \centering
  \includegraphics[width=0.8\textwidth]{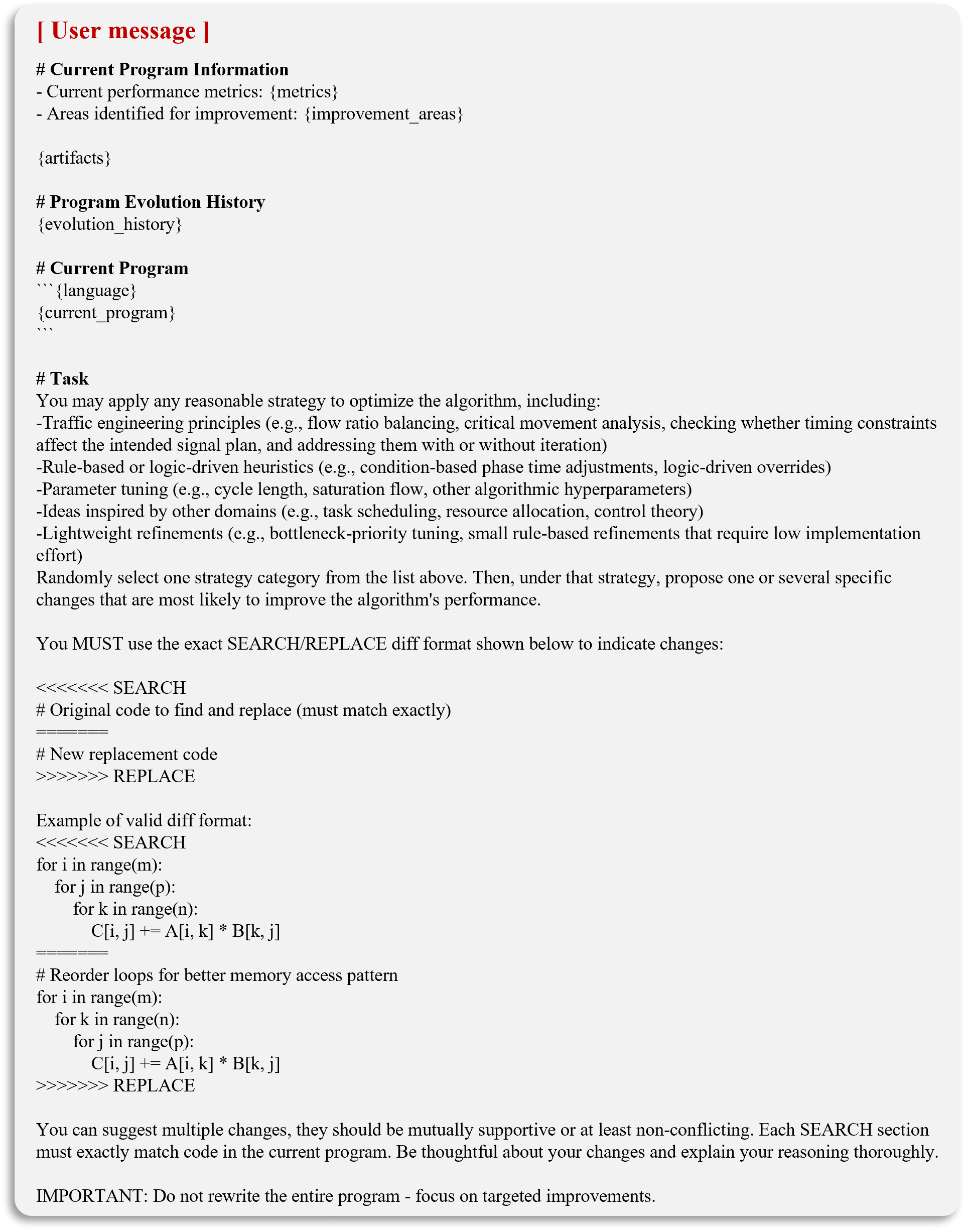}
  \caption{Example user message prompt template}
  \label{fig:user_message}
\end{figure}

\begin{figure}[ht]
  \centering
  \includegraphics[width=0.8\textwidth]{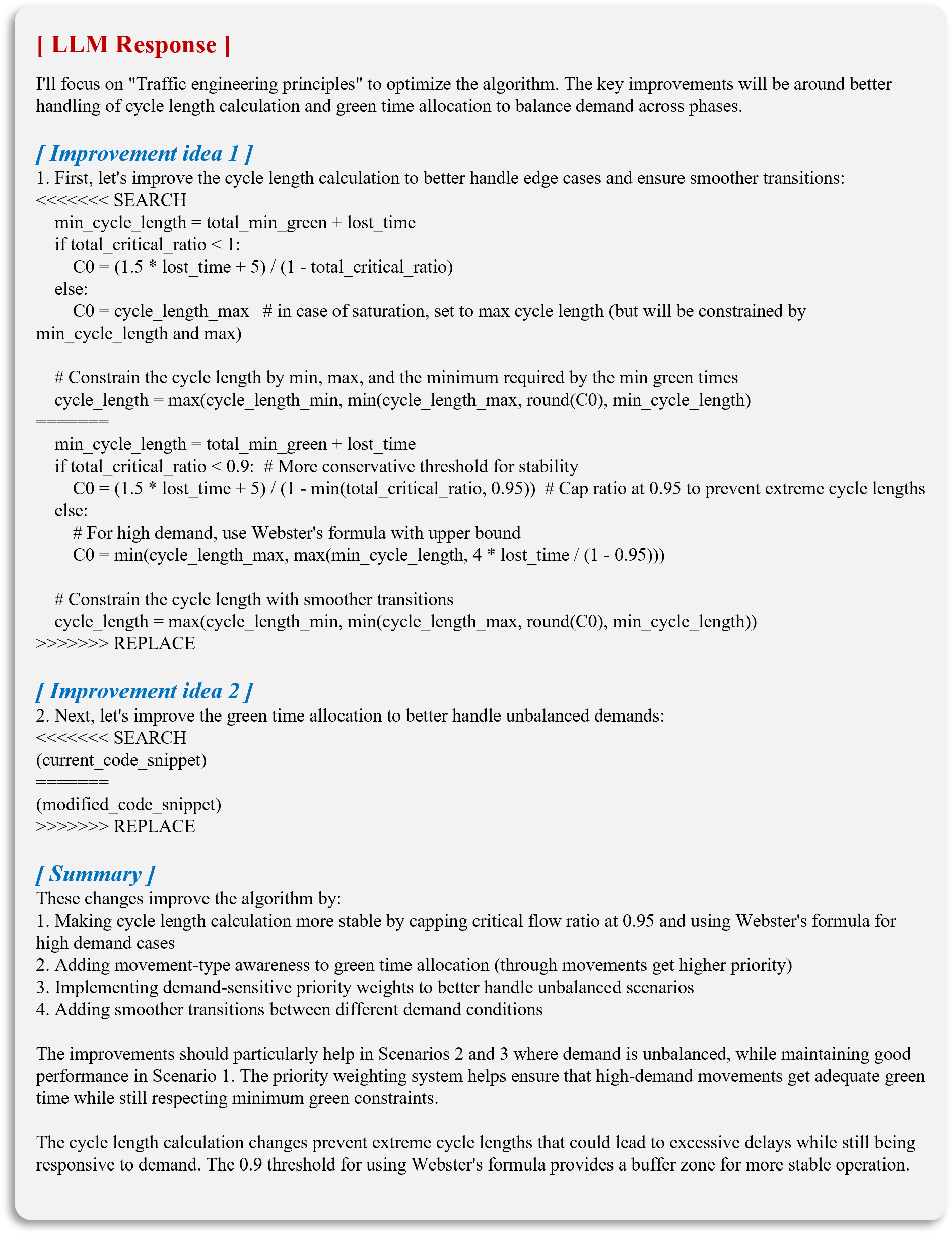}
  \caption{Representative LLM response}
  \label{fig:response}
\end{figure}

\begin{figure}[ht]
  \centering
  \includegraphics[width=0.98\textwidth]{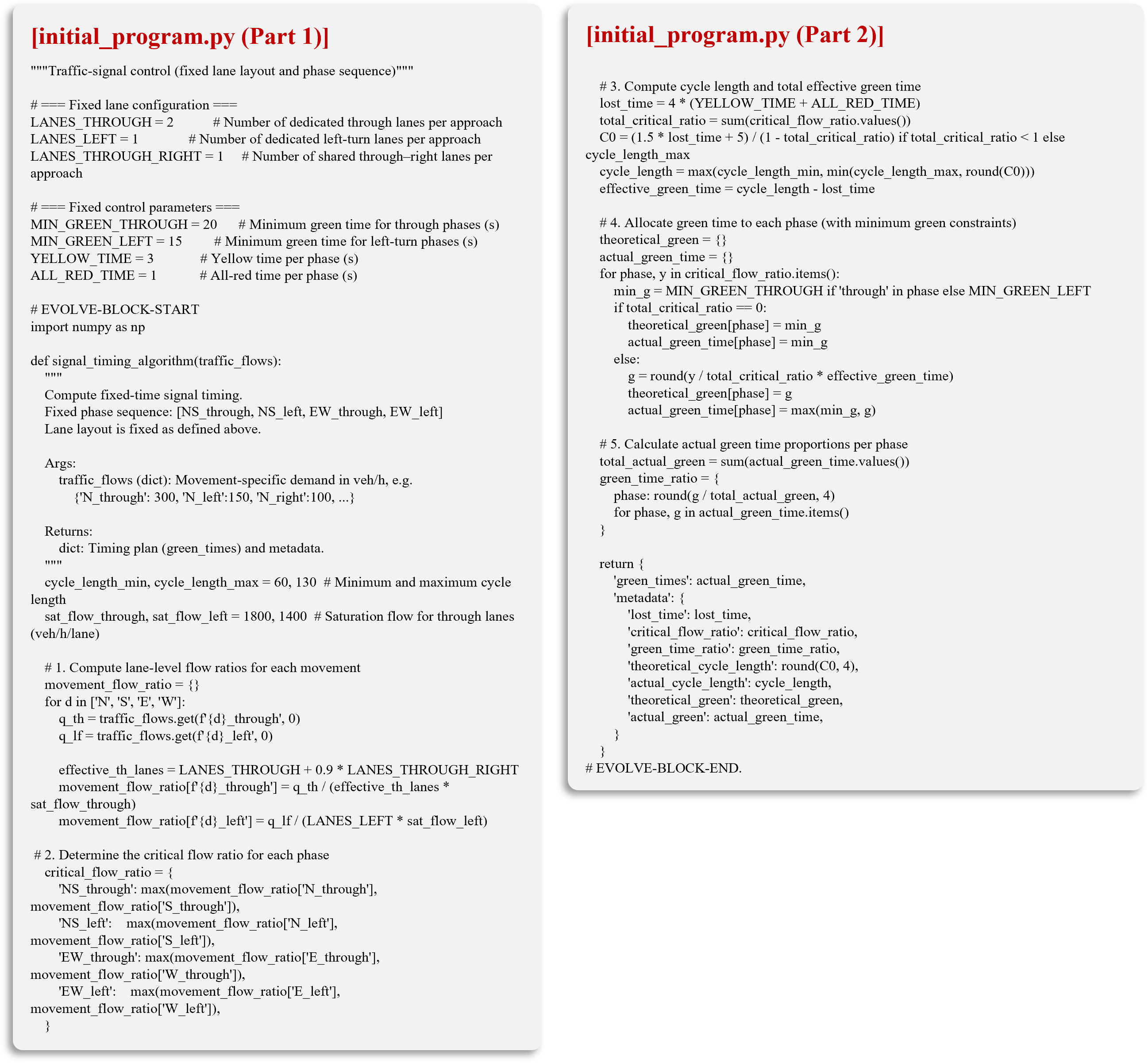}
  \caption{Implementation of the initial program (\texttt{initial\_program.py})}
  \label{fig:initial_program}
\end{figure}

\begin{figure}[ht]
  \centering
  \includegraphics[width=0.98\textwidth]{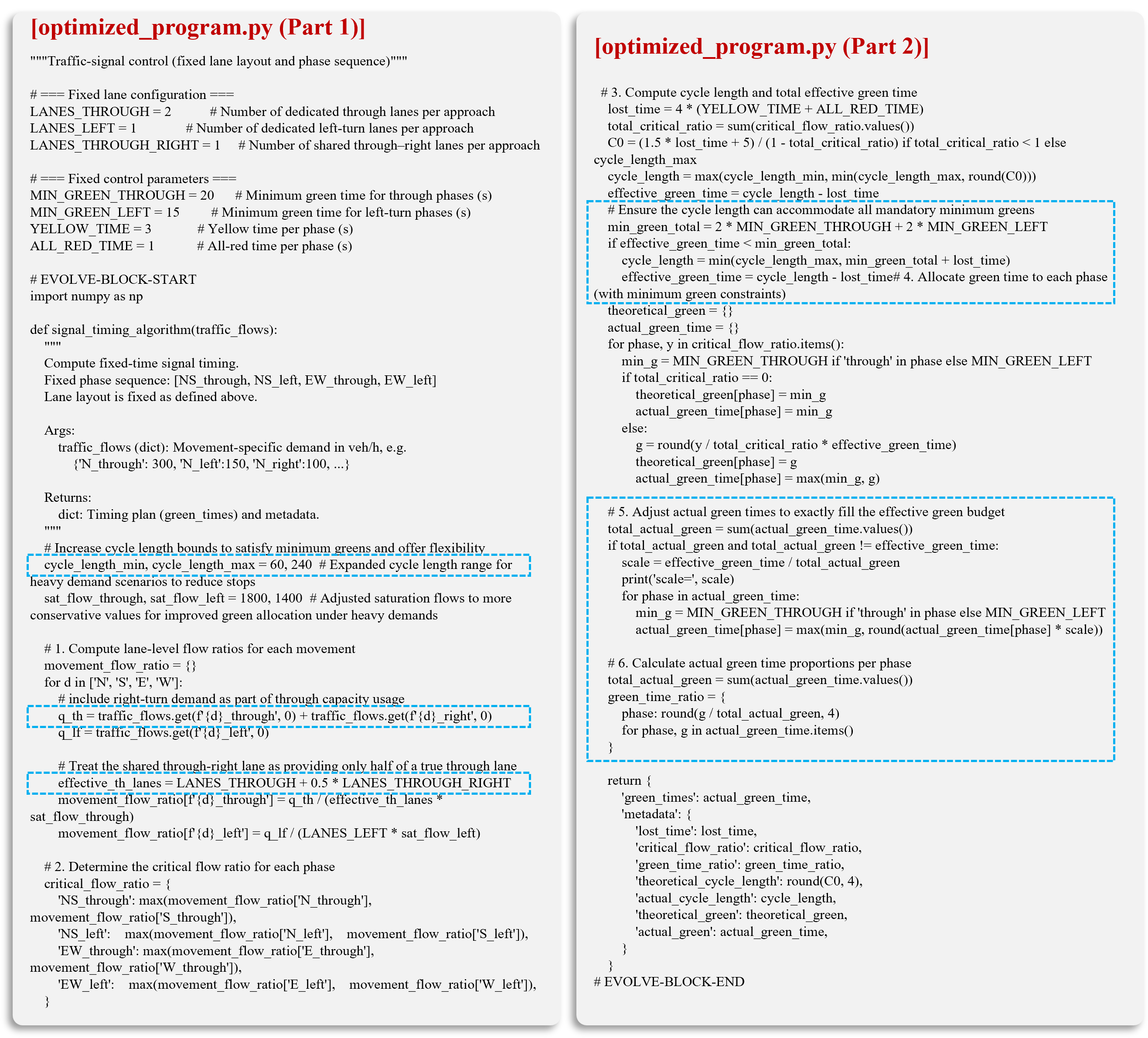}
  \caption{Implementation of the discovered program (\texttt{discovered\_program.py}), with key modifications highlighted (blue dashed lines)}
  \label{fig:optimized_program}
\end{figure}

\begin{figure}[ht]
  \centering
  \includegraphics[width=0.98\textwidth]{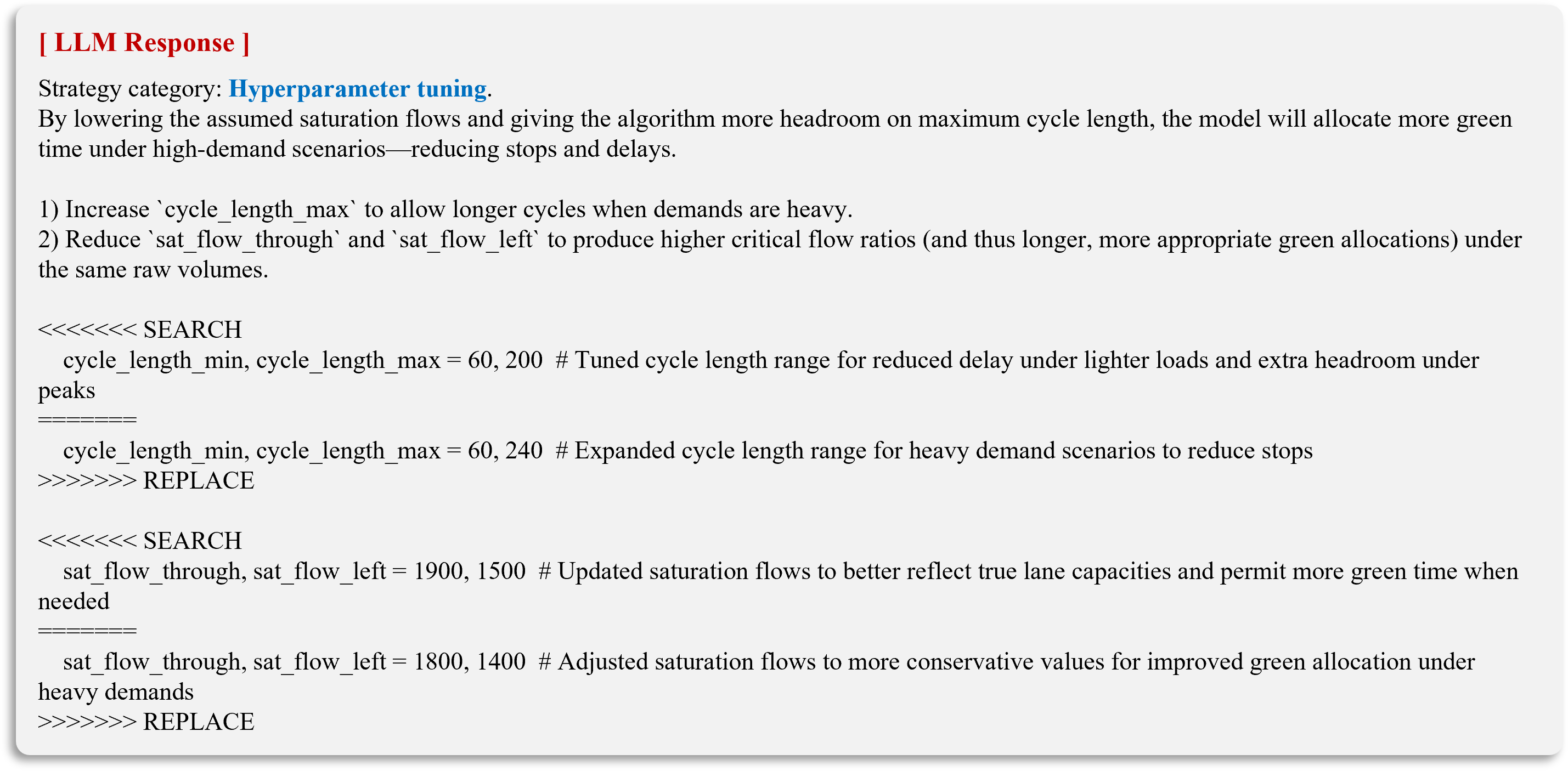}
  \caption{Example LLM response introducing the CLB modification}
  \label{fig:CLB}
\end{figure}

\begin{figure}[ht]
  \centering
  \includegraphics[width=0.98\textwidth]{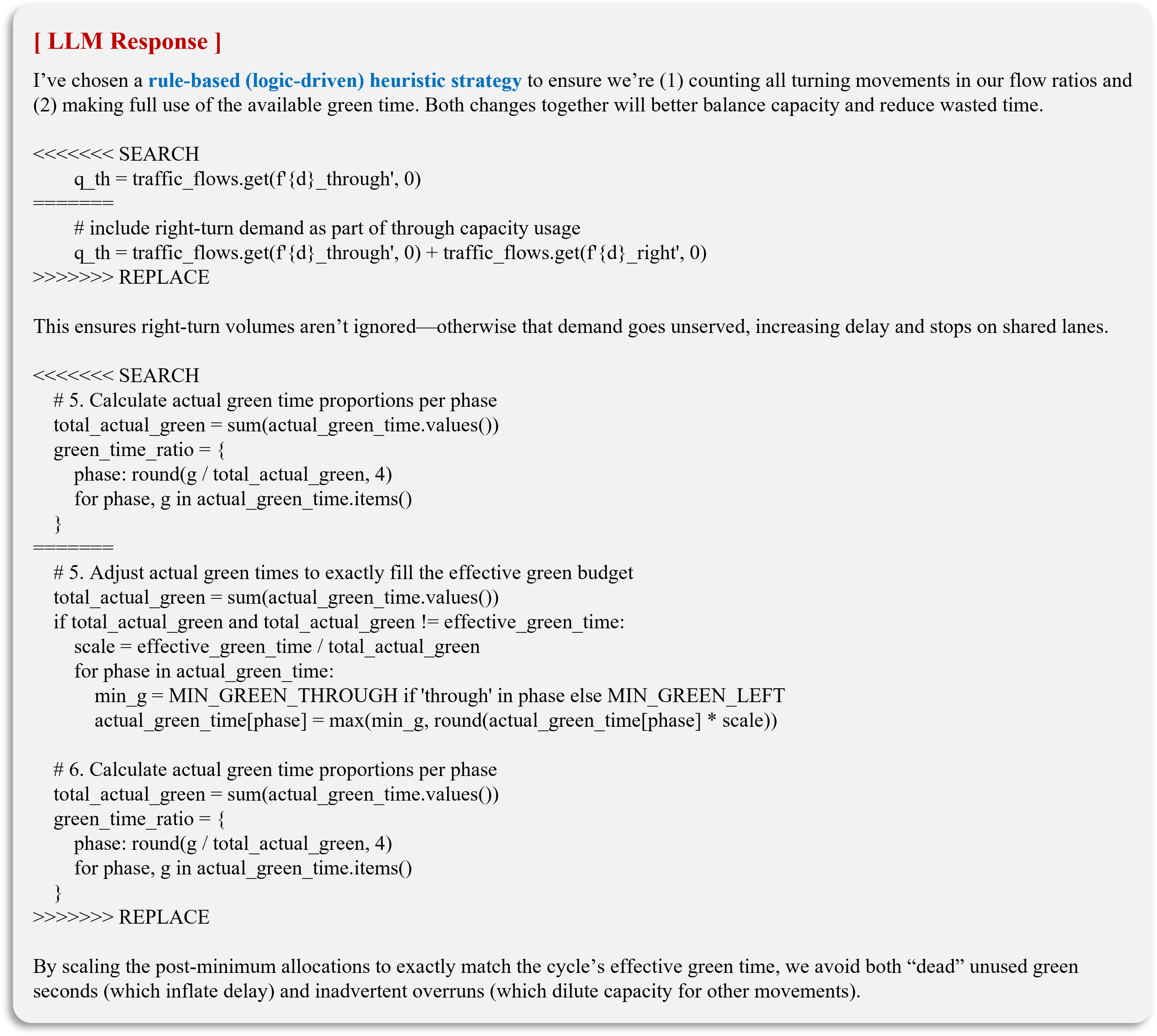}
  \caption{Example LLM response introducing the RTI and PAR modifications}
  \label{fig:RTI_and_PAR}
\end{figure}

\begin{figure}[ht]
  \centering
  \includegraphics[width=0.98\textwidth]{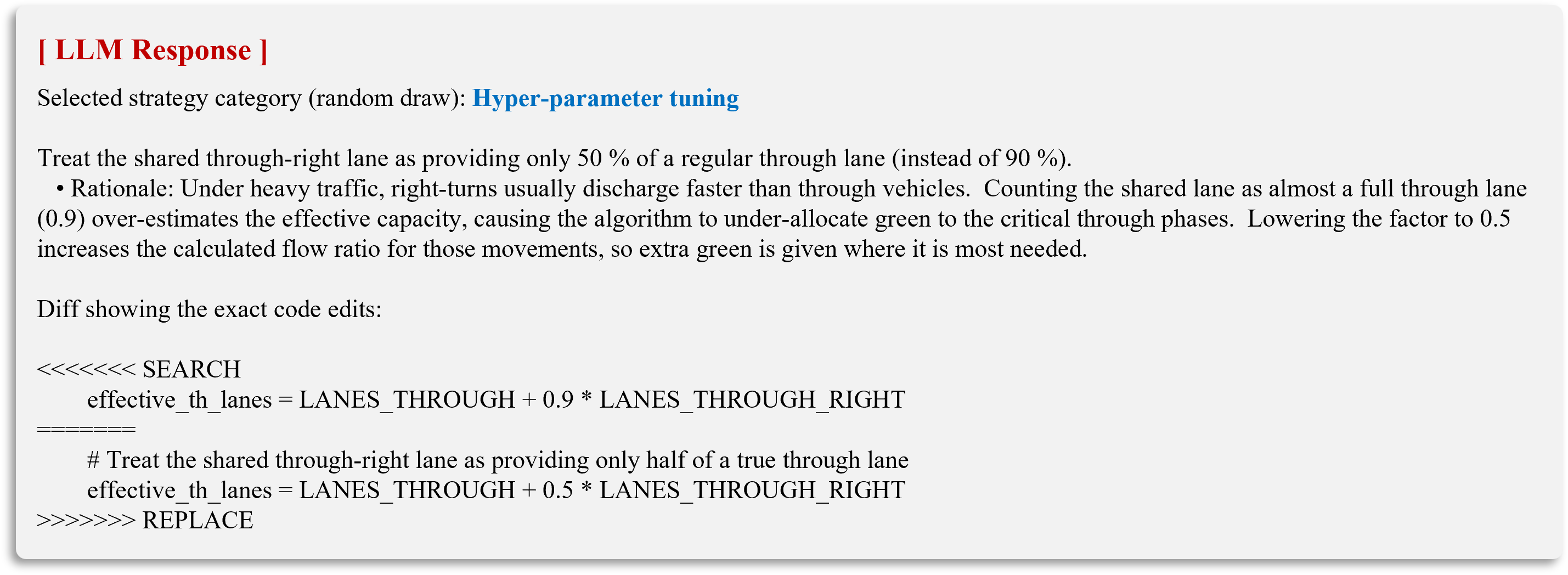}
  \caption{Example LLM response introducing the SLF modification}
  \label{fig:SLF}
\end{figure}

\begin{figure}[ht]
  \centering
  \includegraphics[width=0.98\textwidth]{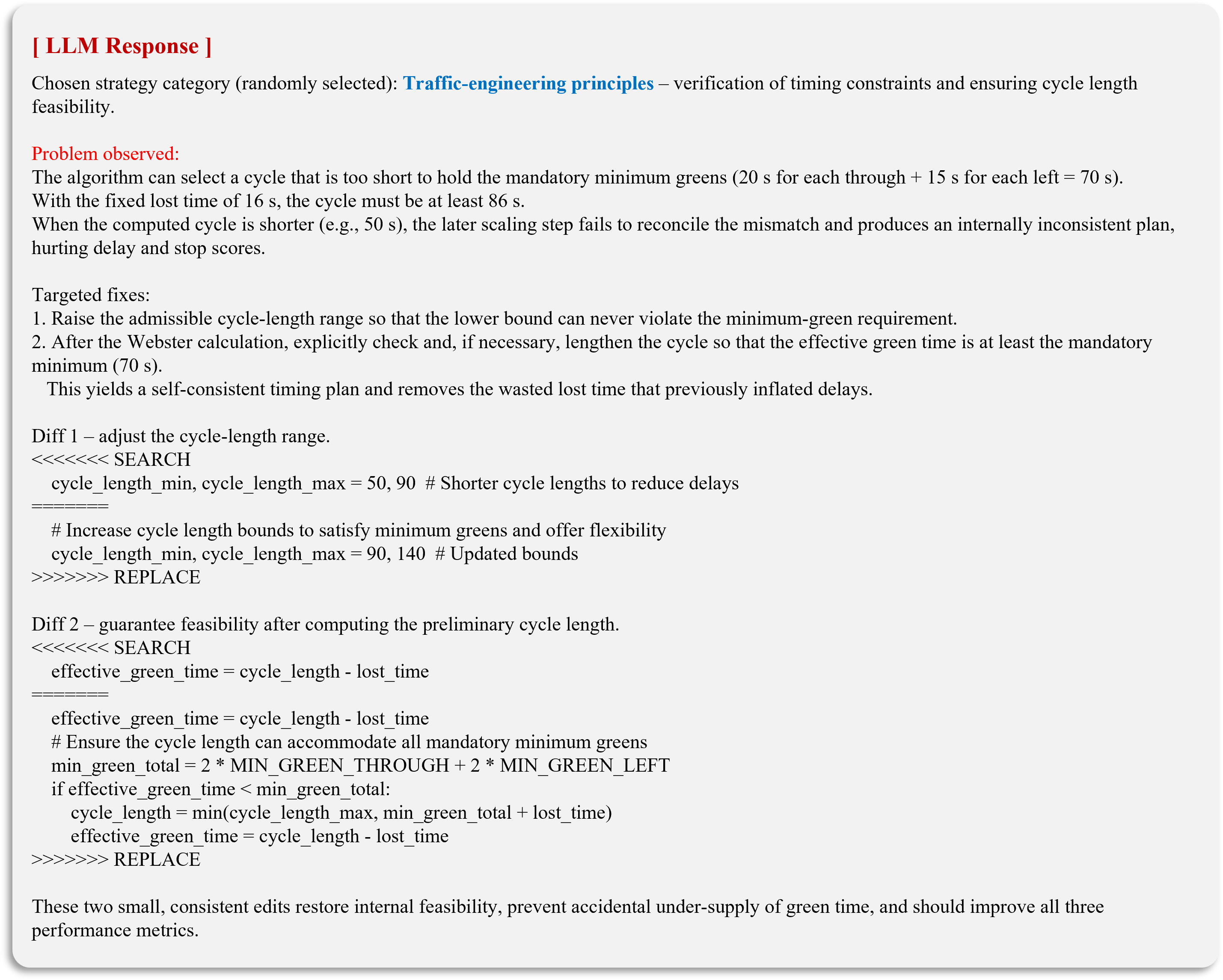}
  \caption{Example LLM response introducing the MGF modification}
  \label{fig:MGF}
\end{figure}

\begin{figure}[ht]
  \centering
  \includegraphics[width=0.8\textwidth]{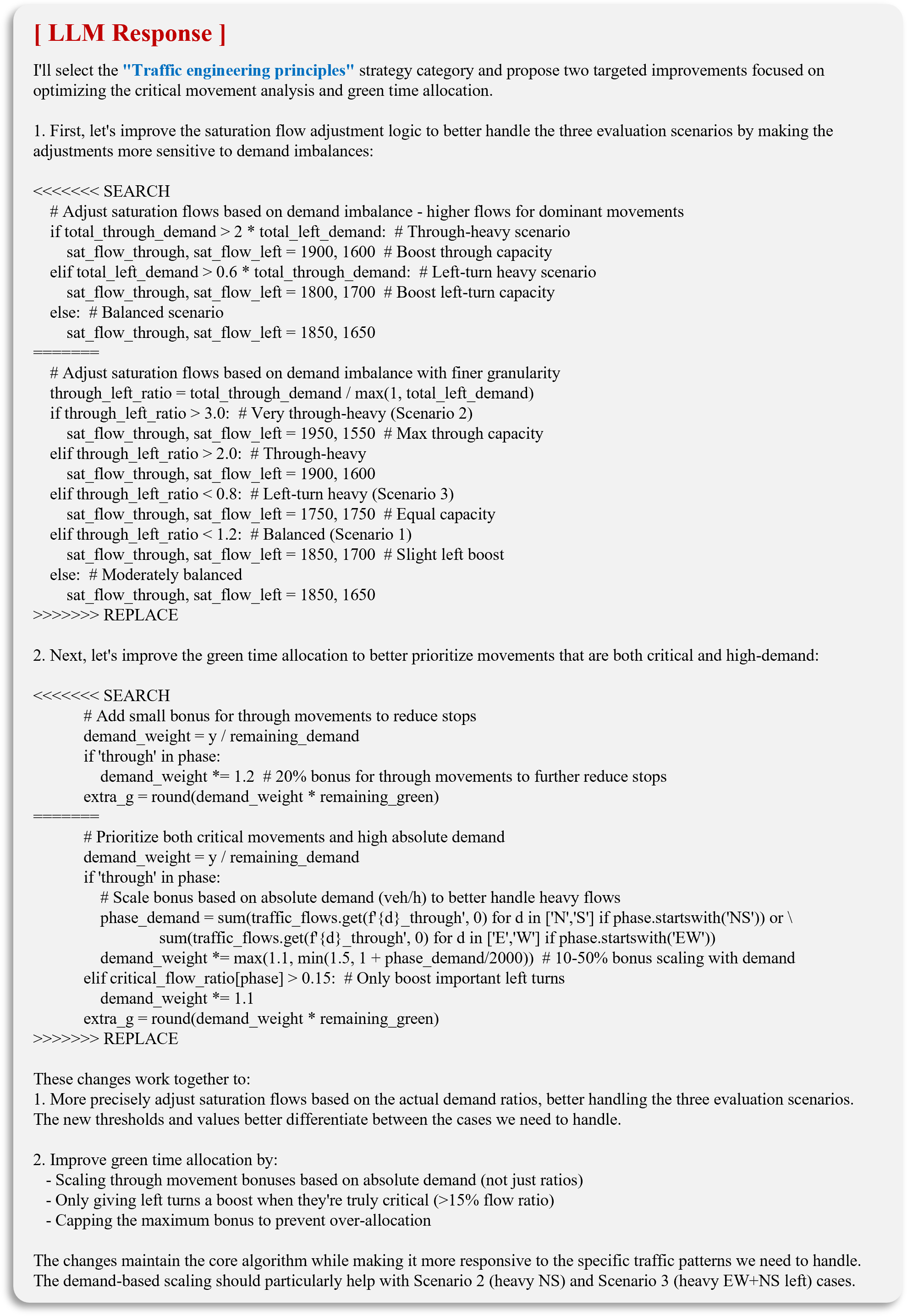}
  \caption{Examples of discarded modifications observed during evolution}
  \label{fig:discarded_modifications}
\end{figure}

\begin{figure}
  \centering
  \includegraphics[width=0.55\textwidth]{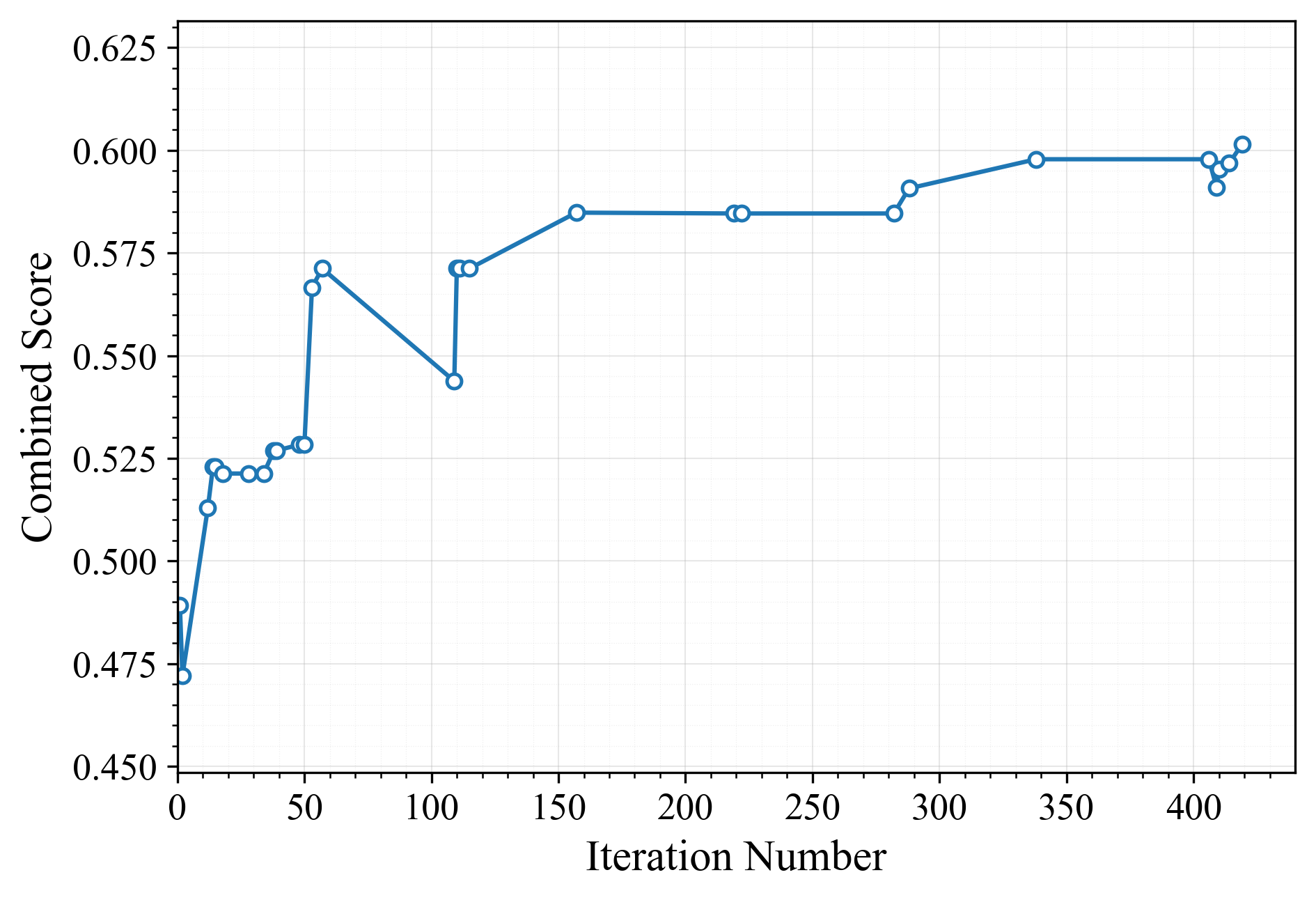}
  \caption{Evolution path of the best discovered program in a 600-iteration run (alternative experiment)}
  \label{fig:evolution600}
\end{figure}

This section provides supplementary examples of prompts and LLM responses used in EvolveSignal. Figs.~\ref{fig:system_message} and \ref{fig:user_message} show representative system and user message templates, while Figure~\ref{fig:response} illustrates a typical LLM response. Figs.~\ref{fig:initial_program} and \ref{fig:optimized_program} present the initial and discovered programs in Python, with key modifications highlighted. Figs.~\ref{fig:CLB}--\ref{fig:MGF} provide detailed examples of individual modifications generated during evolution. Finally, Figs.~\ref{fig:discarded_modifications} and \ref{fig:evolution600} illustrate additional results discussed in Section~\ref{sec:discussion}.

\end{document}